\documentclass{article}

    \PassOptionsToPackage{numbers, compress}{natbib}
\usepackage{paralist}

\usepackage[preprint]{neurips_2024}

\usepackage[utf8]{inputenc} % allow utf-8 input
\usepackage[T1]{fontenc}    % use 8-bit T1 fonts
\usepackage{hyperref}       % hyperlinks
\usepackage{url}            % simple URL typesetting
\usepackage{booktabs}       % professional-quality tables
\usepackage{amsfonts}       % blackboard math symbols
\usepackage{nicefrac}       % compact symbols for 1/2, etc.
\usepackage{microtype}      % microtypography
\usepackage{xcolor}         % colors
\usepackage{graphicx}
\usepackage{caption, subcaption}  %%{subfigure}
\usepackage{multirow}
\usepackage{algorithm}
\usepackage[noend]{algorithmic}
\usepackage{amsmath}
\usepackage{array} % required for text wrapping in tables
\usepackage[most]{tcolorbox}
\usepackage{pifont, arydshln}
\usepackage{hyperref} 

\hypersetup{
    colorlinks,
  citecolor=magenta,
  linkcolor=red,
  urlcolor=blue           % color of external links
}
\usepackage{numprint}

\newcommand{\cmark}{\ding{51}}%
\newcommand{\xmark}{\ding{55}}%

\usepackage{amssymb}
\usepackage{mathtools}
\usepackage{amsthm}
\usepackage{enumitem}
\usepackage[capitalize,noabbrev]{cleveref}
\usepackage{wrapfig}
\usepackage[normalem]{ulem} % for strikeout font
\usepackage{blindtext}%
\usepackage[textsize=tiny]{todonotes}
\presetkeys{todonotes}{fancyline}{}

\newcommand{\ITT}{ITT\xspace}
\newcommand{\TOT}{TOT\xspace}
\newcommand{\dataset}{\textsc{SugarCrepe++} dataset\xspace}
\newcommand{\scd}{\textsc{SugarCrepe}\xspace}
\newcommand{\dset}{\textsc{SugarCrepe++}\xspace}
\definecolor{airforceblue}{rgb}{0.36, 0.54, 0.66}
\newcommand{\graybox}[1]{\textcolor{airforceblue}{#1}}
\title{\dset Dataset: Vision-Language Model Sensitivity to Semantic and Lexical Alterations}

\author{%
  Sri Harsha Dumpala$^{1,2}$\thanks{\ The authors contribute equally to this work.} \quad
  Aman Jaiswal$^{1}$\footnotemark[1] \quad
  Chandramouli Sastry$^{1,2}$ \\
  {\bf Evangelos Milios$^{1}$} \quad
  {\bf Sageev Oore$^{1,2}$} \quad
  {\bf Hassan Sajjad$^{1}$} \\
  $^1$Dalhousie University, Canada.
  $^2$Vector Institute, Canada. \\
}

\begin{document}

\maketitle

\begin{abstract}
  Despite their remarkable successes, state-of-the-art large language models (LLMs), including vision-and-language models (VLMs) and unimodal language models (ULMs), fail to understand precise semantics. For example, semantically equivalent sentences expressed using different lexical compositions elicit diverging representations. The degree of this divergence and its impact on encoded semantics is not very well understood. In this paper, we introduce the \dataset to analyze the sensitivity of VLMs and ULMs to lexical and semantic alterations. Each sample in \dataset consists of an image and a corresponding triplet of captions: a pair of semantically equivalent but lexically different positive captions and one hard negative caption. This poses a 3-way semantic (in)equivalence problem to the language models. We comprehensively evaluate VLMs and ULMs that differ in architecture, pre-training objectives and datasets to benchmark the performance of \dataset. Experimental results highlight the difficulties of VLMs in distinguishing between lexical and semantic variations, particularly in object attributes and spatial relations. Although VLMs with larger pre-training datasets, model sizes, and multiple pre-training objectives achieve better performance on \dset, there is a significant opportunity for improvement. We show that all the models which achieve better performance on compositionality datasets need not perform equally well on \dset, signifying that compositionality alone may not be sufficient for understanding semantic and lexical alterations. The \dataset serves as a new challenge to the vision-and-language community.\footnote{Data and code is available at \url{https://github.com/Sri-Harsha/scpp}}
\end{abstract}

%------------------------------------------------------------
\section{Introduction}
%------------------------------------------------------------

Large language models (LLMs), including vision-and-language models and unimodal language models, have shown tremendous results in solving a majority of vision and natural language processing (NLP) tasks. 
Surprisingly, despite such success, LLMs can exhibit different behaviours for semantically equivalent sentences composed with different syntactic or lexical structures. Previous works have reported such lack of compositional reasoning in both vision-and-language models (e.g., \citep{thrush2022winoground,yuksekgonul2023and,zhao2022vl,ray2023cola,wang2023can}) and unimodal language models (e.g., \citep{krishna2023paraphrasing, p0, p1, p2}). 
For instance, the performance of the state-of-the-art (SOTA) LLMs including GPT-4, Gemini and Llama are sensitive to the prompt formatting~\citep{sclar2023quantifying}. The model editing techniques~\citep{p0,p2} suffer from misfired edits due to the dominance of lexical overlap~\cite{wang2023editingsurvey}. 
Similarly, \citet{p1} demonstrated that safety aligned models can be ``jailbroken'' by simply appending an adversarial suffix causing them to generate objectionable content bypassing all safeguards. 

These observations suggest that language models' perception of semantic similarity crucially depends on the lexical representation of the sentence and calls for a stricter evaluation of semantic text similarity that factors in lexical and syntactic structures. Semantic text similarity is one of the oldest metrics to evaluate language understanding ~\citep{wong-mooney-2006-learning,peng-roth-2016-two,varelas2005semantic,hliaoutakis2006information,hliaoutakis2006medsearch,islam2012text,soto2015similarity} 
and despite recent evidence of lexical sensitivity, large benchmarks (e.g., ~\cite{bigbench, MTEB}) evaluate semantic similarity without explicitly considering the lexical influence. In this work, we aim to address this gap by proposing a dataset to perform joint evaluation of semantic understanding --- through the semantic equivalence detection task (elaborated below) --- and lexical sensitivity in language models. 

Recognizing semantic similarity is often viewed as being fundamental to language understanding, and strong performance on semantic text similarity is often predictive of a language model's performance in various downstream applications \citep{sts17t1} including question-answering, retrieval and summarization. Based on overlap in their meaning, a pair of sentences can be roughly labelled as semantically equivalent, semantically similar or semantically dissimilar. More specifically, semantically equivalent sentences convey the \textit{same} \textit{meaning}, perhaps differing in terms of syntactic and lexical structures. On the other hand, sentences that are not semantically equivalent but describe the \textit{same} \textit{topic} are said to be semantically similar \citep{sts13}. Important examples include MRPC~\citep{dolan2005automatically}, QQP~\citep{qqp_link} and STS~\citep{sts12t6,sts13,sts14t10,sts15t2,sts16t1}: while MRPC and QQP contain binary labels indicating semantic equivalence, STS uses a score between 0 to 5 to indicate the degree of semantic equivalence. 

The timely release of these datasets have fuelled the research and development of improved language models. While these datasets remain relevant even today and are included as part of the challenging GLUE benchmark~\cite{wang-etal-2018-glue}, we aim to improve upon the following aspects to evaluate language understanding through \textbf{\textit{semantic equivalence task under controlled lexical constraints}}:
\begin{compactenum}[(a)]
    \item Varying Definitions of Semantic Equivalence: Two sentences are said to be semantically equivalent if the sentences convey the same meaning and can be inferred from each other (i.e., bidirectional entailment). While MRPC~\citep{dolan2005automatically} aims to evaluate if language models can detect semantic equivalence, it ultimately uses a loose definition of equivalence in that the sentence pairs that convey different information about the same topic are also considered to be semantically equivalent. Different from semantically equivalent sentences, a pair of questions are defined to be semantically equivalent if they have the same answer \citep{rodrigues-etal-2018-semantic} and hence, datasets on semantically equivalent question pairs (e.g., QQP) require additional knowledge beyond language understanding. In contrast, we focus our evaluation on fundamental language understanding ability and evaluate a language model in terms of its ability to recognize semantic equivalence between a pair of sentences.
    \item Lack of Lexical Constraints: While achieving perfect scores on the existing semantic similarity datasets is indeed challenging, trivial baselines using lexical overlap also provide reasonable estimates of semantic similarity (for example, see Figure 2 and Table 6 in \citet{abdalla-etal-2023-makes}). Therefore, the extent to which language models rely upon lexical structure when identifying semantic equivalence and semantic similarity is not clearly known. We are thus motivated to explore a more challenging setting that requires a language model to encode semantics beyond superficial lexical structure. 
\end{compactenum}

Closer to our goal, \citet{hsieh2023sugarcrepe} introduced the challenging \scd dataset to evaluate the ability of vision-language models (VLMs) to identify \textit{one} correct caption in a pair of lexically similar sentences. As an example, given an image, the model may be asked to select the correct caption between ``\textit{A tractor and two boats far from the water}'' (incorrect) and  ``\textit{A tractor and two boats beside the water}''(correct). \citet{hsieh2023sugarcrepe} reported that several VLMs face challenges in selecting the correct caption and attributed the low
%ered 
performance to the text encoder's inability to identify semantic differences in a pair of lexically similar sentences. While this dataset provides a good starting point, it is insufficient for comprehensively evaluating the lexical sensitivity and semantic understanding of a model; the model's understanding of semantic equivalence in the presence of lexical differences remains unclear from such an evaluation. For instance,
% \css{ because ??} 
``\textit{Couple of boats and a tractor located next to the water}'' is semantically identical to the correct caption despite lexical dissimilarities. 
A precise evaluation of lexical influence upon semantic understanding \textit{should} include pairs of semantically-equivalent, semantically-opposite, lexically-similar, and lexically-dissimilar sentences. 
In this work, we target these four cases and define the following research questions:
\begin{compactenum}
    \item How well do LLMs understand the \textit{semantic \textbf{equivalence}} between a pair of sentences given their \textit{syntactic and lexical \textbf{differences}}?
    
    \item How well do LLMs understand the \textit{semantic \textbf{differences}} between a pair of sentences given their \textit{syntactic and lexical \textbf{similarities}}?
    
\end{compactenum}
To that end, we extend \scd to introduce \dataset which additionally contains semantically equivalent sentences that are lexically dissimilar. The answer to these two questions enable us to evaluate the semantic understanding while disentangling the effect of lexical matches between sentences. 

\begin{figure}[!tb]
\hspace{0.82cm}
\vspace{-0.3cm}
    \centering
    \includegraphics[width=1.0\linewidth]{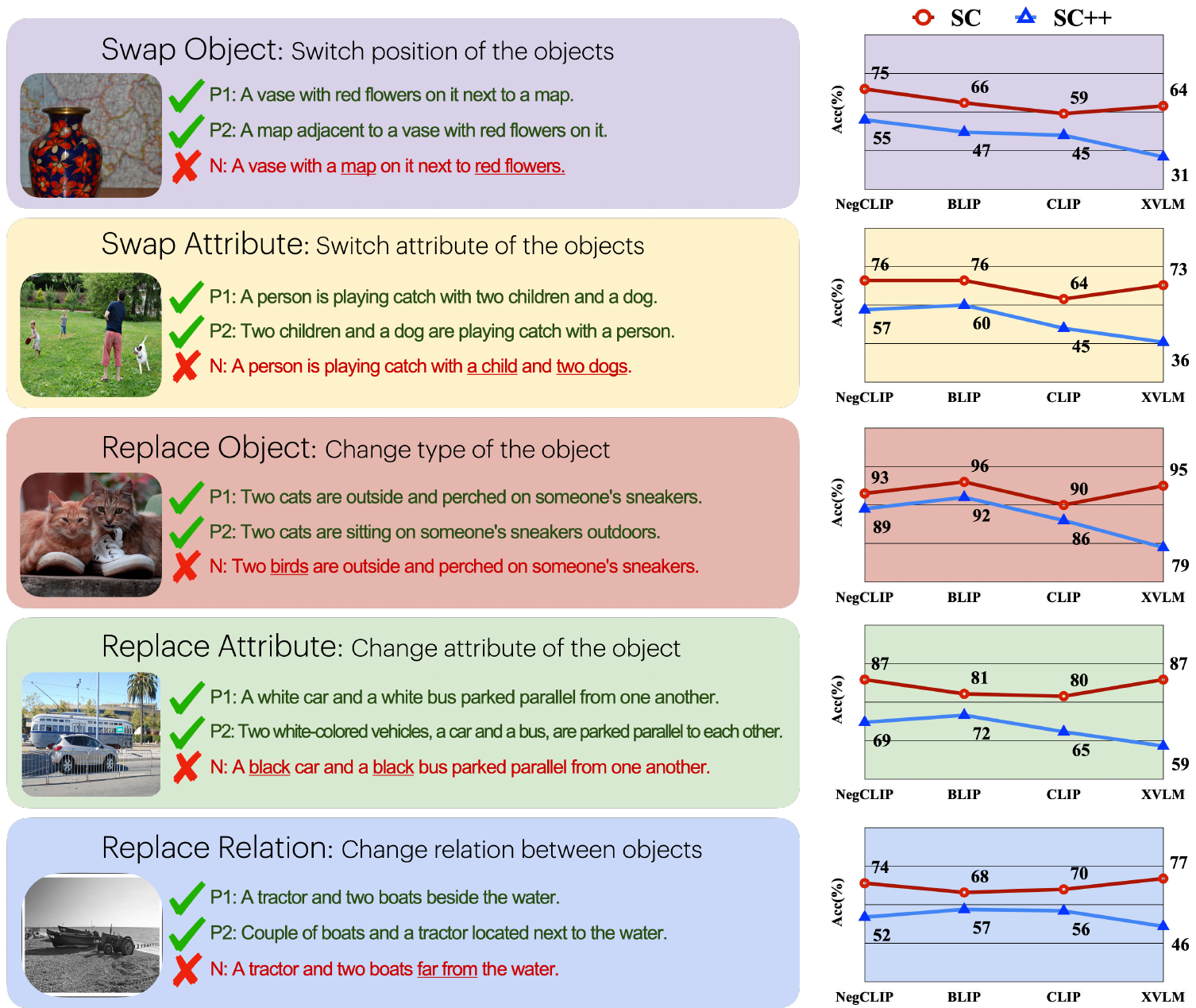}
    \vspace{-8pt}
    \caption{\small 
    %Figure shows an 
    Examples from \dset (SC++) dataset. $P_1$ and $P_2$ are semantically equivalent but lexically different while $N$ is semantically different than both $P_1$ and $P_2$ despite its lexical similarity with $P_1$. The adjacent line charts highlight the performance gaps in VLMs discovered upon re-evaluation using SC++, and shows that strong lexical and semantic understanding may not be required to achieve better performance on \scd (SC).} 
    \label{fig:example}
    \vspace{-8pt}
\end{figure}
 
Our contributions to this work are as follows:
\begin{compactitem}
\item \textbf{\dataset.} We introduce \dset, a diverse, multi-modal and human-validated dataset, especially designed to evaluate the sensitivity of encoded semantics in language models to lexical composition. 
    The introduction of a three-way semantic (in)equivalence task enables evaluation with increased resolution previously not possible with two captions\cite{hsieh2023sugarcrepe}. Figure \ref{fig:example}, illustrates instances from five categories incorporated in \dset and highlights the apparent gaps in performance of VLMs when re-evaluated using \dset. 
  \item \textbf{Unified Evaluation.} We designed \dataset such that the overlap of semantic information between the two positive captions is always higher than between the positive and negative captions, even without considering the image. 
    This allows us to evaluate on the Text-to-Text task twice for the same triplet, using each positive caption as the reference once. These caption triplets combined with an image

enables a previously unexplored dual-mode evaluation of VLMs in both Image-to-Text and Text-to-Text settings.
\end{compactitem}

In this work, we evaluate a comprehensive list of VLMs and standalone/unimodal language models (ULMs) using \dset. 
A few of the notable findings are summarized below:
 
    \begin{compactitem}
    \item VLMs struggle to identify the differences between semantic and lexical alterations, particularly if the lexical alterations are based on swapping attributes or objects or replacing relations. 
    \item There exists a large gap between the VLMs and human-level performance, signifying huge scope for improvement in VLMs.
    \item Text encoders of VLMs form the major bottleneck in achieving better performance on \dset.
    \item Even state-of-the-art ULMs fail to consistently dissociate semantics from lexical forms i.e., they tend to select captions with higher lexical overlap than the ones with higher semantic overlap.
    \end{compactitem}

%------------------------------------------------------------
\section{\dset}
\label{dataset_det}

In this section, we describe the data generation and validation pipeline used to create \dataset. The \scd dataset \cite{hsieh2023sugarcrepe} ---  derived from MS-COCO \cite{lin2014microsoft}, a dataset of image-caption pairs --- consists of a correct caption and an incorrect caption for each image while ensuring that the two captions are lexically similar. To create \dataset, we generate another \textit{correct} caption for each image such that it uses alternative lexical representation while being semantically identical to the original correct caption in \scd. If we refer to the correct captions as positives, the incorrect caption can be termed as a \textit{hard negative} due to its high lexical overlap with one of the positive captions.
In contrast to \scd, \dset enables evaluation across both multimodal and unimodal settings while also providing a comprehensive evaluation of semantic understanding with lexical constraints. Additional related work is discussed in \cref{related_works}.

%%%%%%%%%%%%%%%%%%%%%%%%%%%%%%%%%%%%%%%%%
\subsection{Dataset Generation and Validation}
\label{sec:generation-pipe}
%%%%%%%%%%%%%%%%%%%%%%%%%%%%%%%%%%%%%%%%%
\begin{wrapfigure}[24]{c}{0.55\linewidth}
\vspace{-0.75cm}
    \centering
    %%%%%%%%% Algo 1 %%%%%%%%%%
    \begin{minipage}{0.9\linewidth}
    \begin{algorithm}[H]
    \centering
    \caption{Generation Pipeline}
    \label{alg:dataset_generation}
    \begin{algorithmic}[1]
    \REQUIRE Original caption $P_1$
    \STATE $P_2 \leftarrow LLM(M\{P_1\})$
    \STATE Accept $\leftarrow False$
    \WHILE{\textbf{not} Accept}
        \IF {Duplicated$(P_1, P_2)$} 
            \STATE $P_2 \leftarrow LLM(M\{P_1;P_2\})$  
            % \STATE 
        % \ENDIF
        \ELSIF {Superfluous$(P_2, P_1)$}
            \STATE $P_2 \leftarrow LLM(M\{P_1\})$
        % \STATE \textbf{continue}

        \ELSE
        \STATE $Accept \leftarrow True$
        \ENDIF
    \ENDWHILE
    \RETURN $P_2$
    \end{algorithmic}
    \end{algorithm}
\end{minipage}\\
\begin{minipage}{0.9\linewidth}
    % \vspace{-1.545cm} % height fix
    \begin{algorithm}[H]
    \caption{Automatic Validation Pipeline}
    \label{alg:auto-validation}
    \begin{algorithmic}[1]
    \REQUIRE Original caption $P_1$, \\ Generated caption $P_2$
    \STATE $Valid \leftarrow LLM(V\{P_2,P_1\})$ 
    % \STATE $InitValidity\leftarrow valid$ 
    \WHILE{\textbf{not} Valid}     
        \STATE $P_2 \leftarrow GenerationPipeline(M\{P_1\})$
        \STATE $valid \leftarrow LLM(V\{P_2,P_1\})$  
    \ENDWHILE
    \textbf{Return} $P_2$
    \end{algorithmic}
    \end{algorithm}
\end{minipage}
\end{wrapfigure}
Prior works \cite{VQA,balanced_vqa_v2,Le2024coco,park2023rococo} have extensively used image-caption pairs from MS-COCO; specifically Crepe \cite{ma2023crepe} and its improved derivative \scd
\cite{hsieh2023sugarcrepe} that leverages the recent advancements in conditional text generation using large language models (LLMs) to generate hard negative captions, thereby overcoming the issues with procedurally generated captions. The \scd dataset consists of (only) one positive and one hard negative caption for each image. 

Relative to the negative caption, a single positive caption can either have low or high lexical overlap. The original \scd only captures the high overlap case. To evaluate the sensitivity of encoded semantics to lexical alteration, we require an additional positive caption with a different lexical composition.
% low lexical overlap with the existing positive caption.
We build \dset using instruction fine-tuned Mistral 7B \cite{jiang2023mistral} model to further introduce an additional positive caption
%over five negative caption category types 
as illustrated in \cref{alg:dataset_generation,alg:auto-validation}.
The generation process of the additional caption can be divided into two stages: (1) generation using meta-prompts and (2) automated and human validation of generated captions. \\

\textbf{Generation using meta-prompts:} Algorithm~\ref{alg:dataset_generation} presents our generation pipeline. 
We followed an iterative prompting methodology to refine a meta-prompt (M) that generates optimal second positive captions (P$_2$) given the original positive caption (P$_1$).  The meta-prompt (M) is a composition of constituent prompts (M$_i$) formed by concatenating (;) different sub-prompts. 
$$ M = M_1; M_2; M_3; \dots ; M_i $$
The meta-prompt is applied to an input ($x$) to obtain the final prompt M\{x\} that conditions the LLM to generate the second positive caption (P$_2$) as $LLM(M\{P_1\}) \xrightarrow{} P_2 $. We find integrating different techniques into a larger meta-prompt ($M$) improves the generation quality, exploiting the benefits from each of them. Following \cite{role-playing}, we prefix $M$ with a `role-play' prompt (M$_1$) that conditions the LLM to the high-level task of data generation and simulates the role of a `Data Generating AI.' We utilize `Rules Prompting' \cite{eval-ins-following} to enhance LLMs' faithfulness in instruction following using explicit and itemized rules: in particular, we further condition M using a `rules' prompt (M$_2$) that describes three rules to ensure the consistency of the generated caption (P$_2$). Expanding on the prompting methodology from \cite{hsieh2023sugarcrepe}, we also append few-shot `demonstrations' (M$_3$) that include additional reasoning and the generated caption. The `role-play' (M$_1$) prompt is described in Figure \ref{fig:datagenai-instructions}. The `rules' (M$_2$) and `demonstrations' (M$_3$) sub-prompts are elaborated in Figure \ref{fig:prompting} of Appendix \ref{appendix: benchmark generation}. During our initial testing, we noticed the following systematically recurring errors: (i) LLM generated caption is \textit{identical} to the input caption; (ii) parsing failure due to superfluous outputs. Consequently, we incorporate the following safeguards in the generation pipeline as shown in \cref{alg:dataset_generation}: 1) If the generated caption is found to be identical to the input caption (using automatic tools), we use a meta-prompt that includes the complete context with basic instructions to regenerate P$_2$; 2) We detect superfluous outputs based on word overlap and then discard any generation that does not meet the minimum threshold.

\begin{figure}[tb]
   % \vspace{-0.2cm}
    \centering
    \begin{tcolorbox}[
        colback=gray!10, 
        colframe=gray,
        arc=5mm,
        boxrule=1pt,
    ]
      \small \textbf{Role-playing Prompt:} You are an instruction-following DataGenAI. Your expertise lies in accurately interpreting and generating datasets based on given instructions. Your responses are expected to be precise and limited to the required output.
    \end{tcolorbox}
    \vspace{-7pt}
    \caption{Role playing prompt for ``Data Generator AI".}
    \label{fig:datagenai-instructions}
    \vspace{-10pt}
\end{figure}

\textbf{Automated and Human Validation of Generated Captions: }
Since the above safeguards do not ensure that the generated caption is \textit{always} semantically equivalent to the input caption, we require additional steps to ensure semantic equivalence. We notice subtle differences in generated caption that break the semantic equivalence and this also highlights the limitations of semantic understanding in LLMs. Despite such limitations, prior works \cite{simu,chateval} have demonstrated that LLM agents --- e.g., LLMs instantiated with different prompts --- can interact with each other to solve complex tasks. We build on this to reduce human effort and optimize labelling costs using automatic validation of generated captions similar to ~\cite{llm-judge,eval-ins-following}. We employ a validator LLM agent, that is responsible for validating the semantic consistency of the generated caption with the original caption and signalling the Generator agent to retry as needed. Here, we refer to the LLM agents as two instances of the same LLM conditioned on different meta-prompts.

We define a validation meta-prompt (V) that uses P$_1$ and P$_2$ from the previous step to form a validation input to the validator LLM. Similar to the meta-prompt (M), the validation prompt (V) consists of a sequence of sub-prompts, including the `Validation instruction' (V$_1$) prompt and the `Validation demonstration' (V$_2$). The validator LLM is conditioned using V to generate a boolean value indicating the caption's validity,

 $LLM(V\{P_1,P_2\}) \rightarrow \{True, False\}$,
where $V=V_1; V_2$. We use the boolean output value to trigger a regeneration step that starts the generation pipeline again as described in Algorithm \ref{alg:auto-validation}.

The prompt for employing validator LLM is detailed in Figure \ref{fig:LLM-evaluation} in Appendix \ref{app:val_prompt}. 
%The automatic generation and validation algorithm is described in Figure \ref{alg:dataset_generation}.
%
In our automatic validation, we assume that the image is captioned correctly and do not consider the image when ensuring semantic correctness. 

\begin{table}[!b]
\caption{\dset consists of 4757 examples with the following distribution of sample sizes. We exclude the add-attribute and add-objects subsets of SUGARCREPE as they are not applicable to both the image-text task (ITT) and text-only task (TOT) of SugarCrepe++. This exclusion is necessary because the negative captions in these subsets may require visual information to be accurately identified as negative. }
\label{tab:sc_stats}
\vspace{0.3cm}
    \centering
    \begin{tabular}{cccccc}
    \toprule
    Swap Object & Swap Attribute& Replace Object& Replace Attribute& Replace relation & \textbf{Total}\\
    \midrule

    245& 666& 1652         & 788              & 1406     &  4757\\
    \bottomrule
    \end{tabular}
\end{table}

To ensure the quality of the \dataset, we conducted human validation with two experts to correct the errors in the positive sentences (P$_2$) generated by LLMs (Appendix \ref{err_analysis_llm} provides a list of common errors generated by LLMs) and any disagreements between the expert annotators were mutually resolved through inter-annotator discussion. These human annotators also 
assessed the validity of caption triplets (P$_1$, P$_2$, $N$) and paired images ($I$) as a data point.
The final statistics of the \dataset after the human validation are described in Table \ref{tab:sc_stats}.

%------------------------------------------------------------
\section{Benchmark on \dset Dataset}
%------------------------------------------------------------
\textbf{Experimental Setup}
\label{evaluate}
%------------------------------------------------------------
We benchmark VLM performance on \dataset under two different evaluation settings. (1) Multi-modal image-to-text (\ITT) evaluation: both image and text are provided as inputs to evaluate VLMs in a multi-modal setting. (2) Uni-modal text-only (\TOT) evaluation: only text is provided as input to evaluate the text encoders of VLMs in a unimodal setting.

Each sample in the \dataset consists of an image $I$ and corresponding two positive captions (P$_1$ and P$_2$) and a negative caption (N). 
 
If $p(X|I)$ denotes the likelihood of caption $X$ for image $I$, we compute the \ITT evaluation metric given P$_1$, P$_2$ and N as:

\[ \text{ITT}_{hit} = \begin{cases} 1 & (p(P_1|I) > p(N|I)) \land (p(P_2|I) > p(N|I)) \\
0 & \text{otherwise} \end{cases} \]
For a VLM model such as CLIP which relies upon embeddings, the log-likelihood $\log p$ is defined to be proportional to the cosine similarity between the respective embeddings. 
Similarly, 
%in 
the 
\TOT evaluation is defined as: 
\[ \text{TOT}_{hit} = \begin{cases} 1 & (p(P_1|P_2) > p(N|P_2)) \land (p(P_2|P_1) > p(N|P_1)). \\
0 & \text{otherwise} \end{cases} \]

As above, we use cosine-similarity for embedding-based models. We report the performance in terms of Accuracy (\%), computed as the ratio of the number of hits to the total number of samples.
we also perform a human evaluation to calculate the human performance on the benchmark. Human evaluation is performed by $4$ graduate-level students where each person was provided with a randomly selected $150$ samples ($30$ from each subset) and was asked to select the negative caption for each sample. In \TOT setting, only the three captions were provided. In the \ITT setting, image along with the captions were provided to the human evaluators. The average human performance is reported in terms of accuracy (\%).

\begin{table}[!b]
\vspace{-0.3cm}
\caption{\small Comparison of VLMs performance on \dset. Performance reported in terms of Accuracy (\%). Overall best values are in bold, and group-level best values are underlined.}
\label{vlm_sc_pp}
\vspace{0.2cm}
\centering
\footnotesize
\resizebox{1\linewidth}{!}{
\begin{tabular}{lcccccccccc}
\toprule
Model & \multicolumn{2}{c}{Swap Object} & \multicolumn{2}{c}{Swap Attribute} & \multicolumn{2}{c}{Replace Object} & \multicolumn{2}{c}{Replace Attribute} & \multicolumn{2}{c}{Replace Relation} \\
\cmidrule(lr){2-3} \cmidrule(lr){4-5} \cmidrule(lr){6-7} \cmidrule(lr){8-9} \cmidrule(lr){10-11}
 & \multicolumn{1}{c}{\ITT} & \multicolumn{1}{c}{\TOT} & \multicolumn{1}{c}{\ITT} & \multicolumn{1}{c}{\TOT} & \multicolumn{1}{c}{\ITT} & \multicolumn{1}{c}{\TOT} & \multicolumn{1}{c}{\ITT} & \multicolumn{1}{c}{\TOT} & \multicolumn{1}{c}{\ITT} & \multicolumn{1}{c}{\TOT} \\
 \midrule
 Human & 100.00 & 96.67 & 96.67 & 93.3 & 100.00 & 95.00 & 100.00 & 98.33 & 100.00 & 96.67 \\
 \midrule
CLIP \cite{radford2021learning} & \underline{45.18} & 19.74 & 45.21 & 33.03 & 86.80 & 83.72 & 65.61 & 59.14 & 56.26 & 38.62 \\
RoBERTa-ViT-B/32 \cite{schuhmann2022laion} & 44.30 & \underline{29.39} & \underline{56.32} & \underline{52.66} & 89.04 & \underline{94.55} & \textbf{74.49} & \textbf{80.46} & \underline{59.39} & \underline{57.75} \\
ALIGN  \cite{jia2021scaling} & 41.23 & 25.43 & 51.90 & 41.40 & \underline{90.19} & 84.62 & 69.92 & 69.04 & 51.71 & 45.23 \\
ALIP \cite{yang2023alip} & 36.84 & 20.18 & 46.12 & 28.77 & 71.49 & 50.06 & 54.95 & 34.52 & 47.80 & 23.47 \\
\midrule
FLAVA \cite{singh2022flava} & \textbf{54.39} & \textbf{45.18} & 59.21 & \textbf{57.84} & 89.59 & \underline{94.43} & \underline{73.35} & 72.46 & \textbf{60.10} & \textbf{57.97} \\
ALBEF \cite{li2021albef} & 28.94 & 10.09 & 36.83 & 18.87 & 76.27 & 55.57 & 56.35 & 30.33 & 47.80 & 30.65 \\
BLIP \cite{li2022blip} & 47.37 & 31.14  & \textbf{60.58} & 52.97 & \textbf{92.62} & 89.04 & 72.08 & \underline{75.13} & 56.76 & 57.47 \\
BLIP2 \cite{li2023blip2} & 35.09 & 21.49 & 37.60 & 29.98 & 89.41 & 72.58 & 62.82 & 64.47 & 53.27 & 43.47 \\
\midrule
ViLT \cite{kim2021vilt} & 35.23 & \multicolumn{1}{c}{--} & \underline{52.20} & \multicolumn{1}{c}{--} & 91.10 & \multicolumn{1}{c}{--} & 55.33 & \multicolumn{1}{c}{--} & 37.48 & \multicolumn{1}{c}{--} \\
AltCLIP \cite{ChenLZYW23} & 42.54 & 25.43 & 45.81 & 35.77 & \underline{92.61} & 93.46 & \underline{71.06} & \underline{74.62} & 57.25 & \underline{56.69} \\
SegCLIP \cite{luo2023segclip} & \underline{45.61} & \underline{25.44} & 46.12 & \underline{40.64} & 85.90 & \textbf{95.16} & 62.69 & 67.89 & 54.84 & 41.96 \\
XVLM-4M \cite{zeng2022multi} & 31.14 & 10.96 & 36.52 & 19.48 & 79.42 & 67.07 & 59.39 & 40.74 & 46.23 & 29.23 \\
XVLM-16M \cite{zeng2022multi} & 34.21 & 18.86 & 40.33 & 31.05 & 90.81 & 92.07 & 68.02 & 70.43 & \underline{57.47} & 47.87 \\
\bottomrule
\end{tabular}}
\end{table}
%--------------------------------------------------------------------------

\subsection{Evaluation of VLMs on \dset}
%------------------------------------------------------------

We consider a variety of VLMs  for evaluation using \dset: 1) Models trained with a contrastive learning objective such as CLIP~\cite{radford2021learning}, RoBERTa-ViT-B/32~\cite{schuhmann2022laion}, ALIGN~\cite{jia2021scaling} and ALIP~\cite{yang2023alip}. 
2) Models trained by combining multiple objective functions, such as FLAVA~\cite{singh2022flava},  ALBEF~\cite{li2021albef}, BLIP~\cite{li2022blip} and BLIP-2~\cite{li2023blip2}.
3) Models with a unified encoder for text and images, such as ViLT~\cite{kim2021vilt}, and multi-lingual distilled models like AltCLIP~\cite{ChenLZYW23}; 4) Models that align text with corresponding visual concepts in the image, such as SegCLIP~\cite{luo2023segclip}, and XVLM~\cite{zeng2022multi} - with two variants, XVLM-4M and XVLM-16M. We consider a wide array of VLMs that differ in terms of model architecture, total number of parameters, embedding dimension and pre-training objectives to measure the effect of various training choices on the model's semantic understanding capabilities. For further model details, refer to Appendix \ref{appendix:vlm_eval}.

\textbf{Performance of VLMs on \dset is strongly influenced by the type of hard negative. } The performance of VLMs on different subsets of \dset are provided in Table \ref{vlm_sc_pp}. Swap type hard negatives, which are generated by swapping either Objects or attributes, pose a significant challenge to VLMs, as most achieve very low performance. Failing in examples with simple reordering of words, as in swap subset, highlights a key limitation of VLMs in understanding the structure of the input text.  For replace-type hard negatives (generated by replacing objects, attributes, or relations), VLMs are comparatively better at discerning the negative from the positive caption when the object is replaced. VLMs can also somewhat discern hard negatives from positives when attributes are replaced. However, they struggle when the relation between objects is replaced. Additionally, large performance gaps (ranging from 10\% to 50\%) between the best models and human performance across most subsets signifies opportunity for improvement in VLMs' semantic understanding abilities.  \\

\textbf{Pre-training data size and objective functions affect VLM performance.} Table \ref{vlm_sc_pp} shows that the models trained with multiple objective functions, particularly FLAVA and BLIP, perform better on \dset compared to models trained using contrastive loss function alone. This indicates that the contrastive learning objective alone may not be sufficient for VLMs to effectively learn the semantic relations between text and images. Furthermore, models pre-trained with smaller datasets, such as ALIP, ALBEF and XVLM-4M, have lower performance compared to other models. Interestingly, these observations are consistent across all subsets of \dset.\\

\textbf{Text encoders bottleneck VLM performance on \dset. } All VLMs perform significantly better on the \ITT task compared to the \TOT task on \dset (see Table \ref{vlm_sc_pp}). 
This shows that there is a higher ambiguity in identifying the semantic and lexical alterations using only the text embeddings (\TOT) compared to the case of comparing the text embeddings with the image embeddings (\ITT). This is in agreement with the findings in \cite{kamath2023text}. Additionally, FLAVA, the best performaning model on most of the subsets also achieves a good performance in \TOT setting, further signifying the importance of a strong text encoder in achieving better performance. \\

\begin{table}[!htb]
\vspace{-0.3cm}
\caption{\small Evaluation of VLMs fine-tuned for image-text retrieval task. Performance reported in terms of Accuracy (\%). ITR dataset is the dataset used to fine-tune the model for image-to-text retrieval.}
\label{vlm_itr}
\vspace{0.2cm}
\centering
\footnotesize
\resizebox{1\linewidth}{!}{
\begin{tabular}{lccccccccccc}
\toprule
Model & ITR & \multicolumn{2}{c}{Swap Object} & \multicolumn{2}{c}{Swap Attribute} & \multicolumn{2}{c}{Replace Object} & \multicolumn{2}{c}{Replace Attribute} & \multicolumn{2}{c}{Replace Relation} \\
\cmidrule(lr){3-4} \cmidrule(lr){5-6} \cmidrule(lr){7-8} \cmidrule(lr){9-10} \cmidrule(lr){11-12}
 & Dataset & \multicolumn{1}{c}{\ITT} & \multicolumn{1}{c}{\TOT} & \multicolumn{1}{c}{\ITT} & \multicolumn{1}{c}{\TOT} & \multicolumn{1}{c}{\ITT} & \multicolumn{1}{c}{\TOT} & \multicolumn{1}{c}{\ITT} & \multicolumn{1}{c}{\TOT} & \multicolumn{1}{c}{\ITT} & \multicolumn{1}{c}{\TOT} \\
 \midrule
 ViLT \cite{kim2021vilt} & \multicolumn{1}{c}{--} & 35.23 & \multicolumn{1}{c}{--} & 52.20 & \multicolumn{1}{c}{--} & 91.10 & \multicolumn{1}{c}{--} & 55.33 & \multicolumn{1}{c}{--} & 37.48 & \multicolumn{1}{c}{--} \\
 XVLM-16M \cite{zeng2022multi} & \multicolumn{1}{c}{--} & 34.21 & 18.86 & 40.33 & 31.05 & 90.81 & 92.07 & 68.02 & 70.43 & 57.47 & 47.87 \\
 \midrule
ViLT-ITR-COCO \cite{kim2021vilt} & MS-COCO & \textbf{50.88} & \multicolumn{1}{c}{--} & \textbf{73.36} & \multicolumn{1}{c}{--} & 89.89 & \multicolumn{1}{c}{--} & 71.95 & \multicolumn{1}{c}{--} & 61.24 & \multicolumn{1}{c}{--} \\
XVLM-16M-COCO \cite{zeng2022multi} & MS-COCO & 39.91 & 21.06 & 49.93 & \textbf{51.60} & 91.65 & \textbf{96.79} & \textbf{74.24} & \textbf{83.63} & 63.09 & \textbf{62.87} \\
XVLM-16M-Flickr \cite{zeng2022multi} & Flickr & 45.61 & \textbf{21.49} & 50.53 & 44.44 & \textbf{91.71} & 96.01 & \textbf{74.24} & 81.59 & \textbf{64.01} & 59.89 \\ 
\bottomrule
\end{tabular}}
\end{table}					

\textbf{Fine-tuning VLMs for image-text retrieval improves performance with opportunity for further improvements. }
Table \ref{vlm_itr} provides the performance of VLMs (ViLT and XVLM-16M) fine-tuned for the task of image-to-text retrieval (ITR). While we observe performance improvements on \dset, VLMs still face significant challenges in discerning negative captions from positive ones, particularly for the Swap object and Replace relation subsets. Moreover, there remains a substantial gap between VLM performance and human-level performance. %VLMs fine-tuned for ITR still fail to discern negative captions from positive. 
This indicates that VLMs capable of matching images to corresponding captions do not necessarily learn the intricate details regarding semantic information and lexical variations in the text.

\begin{table}[htb]
\vspace{-0.3cm}
\caption{\small Performance of the methods for improving compositionality of VLMs on \dset. Performance reported in terms of Accuracy (\%). Here, the performance of CLIP is the baseline.}
\label{vlm_comp}
\vspace{0.2cm}
\centering
\footnotesize
\resizebox{1\linewidth}{!}{
\begin{tabular}{lcccccccccc}
\toprule
Model & \multicolumn{2}{c}{Swap Object} & \multicolumn{2}{c}{Swap Attribute} & \multicolumn{2}{c}{Replace Object} & \multicolumn{2}{c}{Replace Attribute} & \multicolumn{2}{c}{Replace Relation} \\
\cmidrule(lr){2-3} \cmidrule(lr){4-5} \cmidrule(lr){6-7} \cmidrule(lr){8-9} \cmidrule(lr){10-11}
 & \multicolumn{1}{c}{\ITT} & \multicolumn{1}{c}{\TOT} & \multicolumn{1}{c}{\ITT} & \multicolumn{1}{c}{\TOT} & \multicolumn{1}{c}{\ITT} & \multicolumn{1}{c}{\TOT} & \multicolumn{1}{c}{\ITT} & \multicolumn{1}{c}{\TOT} & \multicolumn{1}{c}{\ITT} & \multicolumn{1}{c}{\TOT} \\
 \midrule
 Human & 100.00 & 96.67 & 96.67 & 93.3 & 100.00 & 97.00 & 100.00 & 98.33 & 100.00 & 96.67 \\
 \midrule
 CLIP \cite{radford2021learning} & 45.18 & 19.74 & 45.21 & 33.03 & 86.80 & 83.72 & 65.61 & 59.14 & \textbf{56.26} & 38.62 \\
 \midrule
NegCLIP \cite{yuksekgonul2023and} & \textbf{55.25} & \textbf{34.65} & \textbf{57.99} & \textbf{56.47} & \textbf{89.53} & \textbf{94.55} & \textbf{69.41} & \textbf{76.27} & 52.27 & \textbf{51.57} \\
CLIP-SVLC \cite{doveh2023teaching} & 42.98 & 18.86 & 48.40 & 34.56 & 80.93 & 91.56 & 56.98 & 66.88 & 47.30 & 51.28 \\
BLIP-SGVL \cite{HerzigMKAFDG23} & 13.16 & \multicolumn{1}{c}{--} & 38.81 & \multicolumn{1}{c}{--} & 53.75 & \multicolumn{1}{c}{--} & 34.39 & \multicolumn{1}{c}{--} & 30.65 & \multicolumn{1}{c}{--} \\
CyCLIP \cite{goel2022cyclip} & 37.72 & 13.60 & 34.40 & 18.72 & 70.28 & 78.29 & 49.87 & 49.12 & 40.41 & 29.87 \\
\bottomrule
\end{tabular}}
\end{table}

% \subsection{Does improving compositionality improve semantic and lexical understanding?}
% \textbf{Evaluating methods that improve compositionality on \dset. }
\textbf{Compositionality enhancing methods improve performance on \dset by strengthening the VLM text encoder. }
We evaluate recent methods proposed to improve compositionality of VLMs, including NegCLIP \cite{yuksekgonul2023and}, SVLC \cite{doveh2023teaching}, CyCLIP \cite{goel2022cyclip}, and BLIP-SGVL \cite{HerzigMKAFDG23}. As shown in Table \ref{vlm_comp}, methods that improve compositionality of CLIP such as NegCLIP and CLIP-SVLC also achieve better performance on \dset compared to CLIP, underscoring the importance of compositionality as a critical component for understanding semantic and lexical differences. Interestingly, the text encoders of models with improved compositionality (NegCLIP and CLIP-SVLC) perform significantly better than the text encoder of CLIP in the \TOT setting. Improved text encoders, in turn, lead to improvements in \ITT. On the other hand, methods such as BLIP-SGVL and CyCLIP, which do not use explicit techniques to strengthen the text encoders, show degradation in performance on \dset. This further highlights the importance of the text encoder in achieving better performance on \dset.

\begin{table}[htb]
\vspace{-0.3cm}
\caption{\small Evaluation of several variants of CLIP on \dset. Performance reported in terms of Accuracy (\%). Best performances in bold. RoB refers to RoBERTa.}
\label{clip_var_sc_pp}
\vspace{0.2cm}
\footnotesize
\centering
\resizebox{\linewidth}{!}{
\begin{tabular}{lrrcccccccccc}
\toprule
 & \#Model & Pre-train & \multicolumn{2}{c}{Swap Object} & \multicolumn{2}{c}{Swap Attribute} & \multicolumn{2}{c}{Replace Object} & \multicolumn{2}{c}{Replace Attribute} & \multicolumn{2}{c}{Replace Relation} \\
\cmidrule(lr){4-5} \cmidrule(lr){6-7} \cmidrule(lr){8-9} \cmidrule(lr){10-11} \cmidrule(lr){12-13}
Model & Params & Data Size  & \multicolumn{1}{c}{\ITT} & \multicolumn{1}{c}{\TOT} & \multicolumn{1}{c}{\ITT} & \multicolumn{1}{c}{\TOT} & \multicolumn{1}{c}{\ITT} & \multicolumn{1}{c}{\TOT} & \multicolumn{1}{c}{\ITT} & \multicolumn{1}{c}{\TOT} & \multicolumn{1}{c}{\ITT} & \multicolumn{1}{c}{\TOT} \\
 \midrule

CLIP \cite{radford2021learning} & 151M & 400M & 45.18 & 19.74 & 45.21 & 33.03 & 86.80 & 83.72 & 65.61 & 59.14 & 56.26 & 38.62 \\
RN50$\times$4 \cite{radford2021learning} & 178M & 400M & \textbf{46.93} & 21.49 & 46.42 & 30.59 & 87.77 & 80.87 & 67.51 & 53.93 & 53.91 & 38.55 \\

RN50$\times$64 \cite{radford2021learning} & 623M & 400M & 44.74 & 16.67 & 45.36 & 31.51 & 90.79 & 73.31 & 64.47 & 48.61 & 54.27 & 38.12 \\
% \midrule
RoB-ViT-B/32 \cite{schuhmann2022laion} & 212M & 2000M & 44.30 & 29.39 & 56.32 & 52.66 & 89.04 & 94.55 & 74.49 & 80.46 & 59.39 & 57.75 \\
ViT-H/14 \cite{schuhmann2022laion} & 986M & 2000M & 43.42 & 27.63 & 54.19 & 50.69 & 93.71 & 90.43 & 71.06 & 73.98 & 56.62 & 51.92 \\
ViT-bigG/14 \cite{schuhmann2022laion} & 2540M & 2000M & 45.61 & 29.82 & \textbf{57.38} & 52.05 & \textbf{94.13} & 90.44 & \textbf{76.41} & 72.84 & \textbf{59.45} & 53.49 \\
XLM-RoB-ViT-B/32 \cite{schuhmann2022laion} & 366M & 5000M & 42.55 & \textbf{30.26} & 55.25 & \textbf{55.56} & 89.41 & \textbf{95.34} & 72.97 & \textbf{80.96} & 55.48 & \textbf{57.82} \\

\bottomrule
\end{tabular}}
\end{table}
%------------------------------------------------------------
%------------------------------------------------------------
\textbf{Larger pre-training data and model size improve CLIP's performance. }
We evaluated variants of CLIP that differ in model architecture and size, as well as pre-training data size, on \dset (see Table \ref{clip_var_sc_pp}). For CLIP variants pre-trained on a dataset of 400 million image-text pairs, smaller models (CLIP and RN50$\times$4) performed better than larger models (RN50$\times$64). Interestingly, larger models (ViT-bigG/14) performed better than smaller models when the pre-training data was increased to 2 billion samples. Moreover, the text encoders also performed better when the pre-training data was increased to 2 billion image-text pairs.

\begin{table}[!tb]
\vspace{-0.3cm}
\caption{\small Comparing the performance of VLMs on \scd and \dset. $\textcolor{green}\uparrow$ and $\textcolor{red}\downarrow$ show increases and decreases in performance with the corresponding CLIP performance as the baseline.}
\label{vlm_sc_scpp}
\vspace{0.2cm}
\centering
\footnotesize
\resizebox{1\linewidth}{!}{
\begin{tabular}{lcccccccccc}
\toprule
Model & \multicolumn{2}{c}{Swap Object} & \multicolumn{2}{c}{Swap Attribute} & \multicolumn{2}{c}{Replace Object} & \multicolumn{2}{c}{Replace Attribute} & \multicolumn{2}{c}{Replace Relation} \\
\cmidrule(lr){2-3} \cmidrule(lr){4-5} \cmidrule(lr){6-7} \cmidrule(lr){8-9} \cmidrule(lr){10-11}
 & \multicolumn{1}{c}{SC} & \multicolumn{1}{c}{SC++} & \multicolumn{1}{c}{SC} & \multicolumn{1}{c}{SC++} & \multicolumn{1}{c}{SC} & \multicolumn{1}{c}{SC++} & \multicolumn{1}{c}{SC} & \multicolumn{1}{c}{SC++} & \multicolumn{1}{c}{SC} & \multicolumn{1}{c}{SC++} \\
 \midrule
 CLIP \cite{radford2021learning} & 59.21 & 45.18 & 64.99 & 45.21 & 90.86 & 86.8  & 80.33 & 65.61 & 70.48 & 56.26 \\
 \midrule
 ALBEF \cite{li2021albef} & 63.16 ($\textcolor{green}\uparrow$) & 28.94($\textcolor{red}\downarrow$) & 69.25($\textcolor{green}\uparrow$) & 36.83($\textcolor{red}\downarrow$) & 93.04($\textcolor{green}\uparrow$) & 76.27($\textcolor{red}\downarrow$) & 84.65($\textcolor{green}\uparrow$) & 56.35($\textcolor{red}\downarrow$) & \textbf{77.60}($\textcolor{green}\uparrow$)  & 47.80($\textcolor{red}\downarrow$)  \\
 XVLM \cite{zeng2022multi} & 64.91($\textcolor{green}\uparrow$) & 31.14($\textcolor{red}\downarrow$) & 73.97($\textcolor{green}\uparrow$) & 36.52($\textcolor{red}\downarrow$) & 95.22($\textcolor{green}\uparrow$) & 79.42($\textcolor{red}\downarrow$) & \textbf{87.69}($\textcolor{green}\uparrow$) & 59.39($\textcolor{red}\downarrow$) & 77.45($\textcolor{green}\uparrow$) & 46.23($\textcolor{red}\downarrow$) \\
 % \hdashline
 BLIP \cite{li2022blip} & 66.22($\textcolor{green}\uparrow$) & 47.37($\textcolor{green}\uparrow$) & 76.25($\textcolor{green}\uparrow$) & \textbf{60.58}($\textcolor{green}\uparrow$) & \textbf{96.55}($\textcolor{green}\uparrow$) & \textbf{92.62}($\textcolor{green}\uparrow$) & 81.98($\textcolor{green}\uparrow$) & \textbf{72.08}($\textcolor{green}\uparrow$) & 68.35($\textcolor{red}\downarrow$) & \textbf{56.76}($\textcolor{green}\uparrow$) \\
 NegCLIP \cite{yuksekgonul2023and} & \textbf{75.44}($\textcolor{green}\uparrow$) & \textbf{55.25}($\textcolor{green}\uparrow$) & \textbf{76.87}($\textcolor{green}\uparrow$) & 57.99($\textcolor{green}\uparrow$) & 93.88($\textcolor{green}\uparrow$) & 89.53($\textcolor{green}\uparrow$) & 87.18($\textcolor{green}\uparrow$) & 69.41($\textcolor{green}\uparrow$) & 74.47($\textcolor{green}\uparrow$) & 52.27($\textcolor{red}\downarrow$) \\
\bottomrule
\end{tabular}}
\end{table}

\textbf{Comparison of performance between \scd and \dset reveals significant differences. }
Table \ref{vlm_sc_scpp} compares the performance of VLMs between \scd and \dset. We observe significant differences in the trend between model performances.
%are observed. 
When using CLIP's performance as the baseline, models such as ALBEF and XVLM which achieve better performance on \scd show significant degradation in performance on \dset. On the other side, there are models such as BLIP and NegCLIP that show improvements on both \scd and \dset. Interestingly for the replace relation subset, NegCLIP achieves better performance on \scd but shows degradation in performance on \dset. These results emphasize the importance of our dataset in evaluating the models in terms of their semantic and lexical understanding, which is not evident by evaluating on compositionality datasets such as \scd.

%------------------------------------------------------------
\subsection{Evaluation of ULMs on \dset }
% 4 3 8
% 9 5 1     
% 2 7 6

We evaluate a comprehensive list of unimodal language models (ULMs) to determine the semantic and lexical sensitivity of text-only models. We sample ULMs covering various model sizes, architectures, and training objectives. Recent state-of-the-art small-sized models include MiniLM \cite{sbert}, GTE \cite{li2023general} and BGE \cite{bge_embedding}. From the results presented in Table \ref{tab:unimodal-performance-summary}, we observe trends similar to the VLM's performance in TOT task but more importantly, we observe ULMs achieve significant improvements on average as compared to VLMs. Nevertheless, we notice that some of these performances are far below human performance --- for instance, we observe a performance gap of $\approx$40\% in swap object and swap attribute subsets. 
Comprehensive results are available in Appendix \ref{appendix:addtional-analysis-ULMs}.

\begin{table*}[!htb]
    \centering
    \caption{\small Comparison of SOTA unimodal language models (ULMs) on \dset. We report the \TOT accuracy  (\%), and group the results row-wise based on the model size as reflected by the parameter count. We include the number of parameters in text encoders relative to BERT-base, i.e., 109.5 Million parameters. Overall, best values are in bold, and group-level best values are underlined. We report the average across different subsets as an additional column. Refer to \cref{tab:unimodal-performance} for additional results.} 
        \label{tab:unimodal-performance-summary}
    \resizebox{\linewidth}{!}{
    \begin{tabular}{@{}lccccccc@{}}
\toprule
Model &  \begin{tabular}[c]{@{}c@{}}\#Params \\ (BERT Scale)\end{tabular} &\begin{tabular}[c]{@{}c@{}}Swap \\ Object\end{tabular} & \begin{tabular}[c]{@{}c@{}}Swap \\ Attribute\end{tabular} & \begin{tabular}[c]{@{}c@{}}Replace \\ Object\end{tabular} & \begin{tabular}[c]{@{}c@{}}Replace \\ Attribute\end{tabular} & \begin{tabular}[c]{@{}c@{}}Replace \\ Relation\end{tabular} & Average \\ 
            \midrule
            Human & & 96.67 & 93.3 & 97.00 &98.33 & 96.67 & 96.40 \\
            \midrule
            BGE-small-en-v1.5 \cite{bge_embedding}&  0.3&15.51& 24.02& 94.19& \underline{75.00}&\underline{75.53}&\underline{56.85}\\
           All-MiniLM-L12-v2 \cite{wang2020minilm} &  0.3&\underline{18.78}& \underline{25.38}& \underline{95.22}& 73.86&70.41&56.73\\
            \midrule 
            Angle-BERT-base \cite{li2023angle}  &  1&\underline{25.71}& \underline{33.63}& 92.07& 78.43&75.32&\underline{61.03}\\
            BGE-base-en-v1.5  \cite{bge_embedding} &  1&17.14& 25.23& \underline{93.83}& \underline{78.55}&\underline{76.10}&58.17\\
            \midrule
            UAE-Large-v1 \cite{li2023angle} &3 & 40.41& 41.44& \textbf{96.85}& \underline{76.14}&\underline{75.82}&66.13\\
            All-RoBERTa-large-v1 \cite{sbert}  &  3.1&42.04& 45.20& 94.61& 74.75&74.96&66.31\\
            Sentence-T5-xl \cite{sent-t5} &  11.3&\textbf{47.35}& \textbf{49.25}& 90.98& 75.38&75.32&\underline{67.66}\\
            \midrule 
             Angle-Llama-7b-nli-v2 \cite{li2023angle}&  62.3&\underline{37.96}& \underline{45.80}& 95.22& 84.39&\textbf{81.44}&\textbf{68.96}\\
            E5-Mistral-7b-instruct \cite{wang2023improving, wang2022text} &  64.9&33.47& 37.84& \underline{96.67}& \textbf{87.06}&80.51&67.11\\
            \bottomrule
        \end{tabular}
        }
\end{table*}

%------------------------------------------------------------
\section{Conclusion}
We introduce \dset, a dataset for evaluating the ability of language models, including both vision-language models (VLMs) and unimodal language models (ULMs), to understand their sensitivity to semantic and lexical alterations in text. Our dataset supports evaluations in both image-to-text (\ITT) and text-to-text (\TOT) settings. We evaluated a comprehensive list of VLMs and ULMs to highlight a fundamental limitation with these language models in understanding semantic and lexical alterations. The key findings from our evaluation are: (1) There is a significant performance gap between VLMs and human-level performance, signifying a huge scope for improvement in VLMs. (2) All VLMs exhibit difficulty in comprehending semantic and lexical alterations, especially when these alterations involve swapping attributes or objects or replacing relations. (3) Similarly, state-of-the-art (SOTA) ULMs lack a robust understanding of lexical composition and consistently fail to separate semantics from lexical forms. (4) While increasing pre-training data, model size, and improving compositionality enhance performance on \dataset, these models still fall considerably short of human performance.  These insights underscore the critical need for advancements to close the performance gap between models and human understanding. Our \dataset serves as a valuable tool for driving future research in this area.

%------------------------------------------------------------
\bibliographystyle{abbrvnat}
\bibliography{references,refs-eem}

%%%%%%%%%%%%%%%%%%%%%%%%%%%%%%%%%%%%%%%%%%%%%%%%%%%%%%%%%%%%
\appendix
\newpage
\section*{Checklist}

\begin{enumerate}

\item For all authors...
\begin{enumerate}
  \item Do the main claims made in the abstract and introduction accurately reflect the paper's contributions and scope?
    \answerYes{}
  \item Did you describe the limitations of your work?
    \answerYes{}
  \item Did you discuss any potential negative societal impacts of your work?
    \answerYes{}
  \item Have you read the ethics review guidelines and ensured that your paper conforms to them?
    \answerYes{}
\end{enumerate}

\item If you are including theoretical results...
\begin{enumerate}
  \item Did you state the full set of assumptions of all theoretical results?
    \answerNA{}
	\item Did you include complete proofs of all theoretical results?
    \answerNA{}
\end{enumerate}

\item If you ran experiments (e.g. for benchmarks)...
\begin{enumerate}
  \item Did you include the code, data, and instructions needed to reproduce the main experimental results (either in the supplemental material or as a URL)?
    \answerYes{\bf We include the Github link to our data and code in Appendix \ref{reproduce}.}
  \item Did you specify all the training details (e.g., data splits, hyperparameters, how they were chosen)?
    \answerNA{\bf We perform zero-shot evaluation of models. We did not train any models for this work.}
	\item Did you report error bars (e.g., with respect to the random seed after running experiments multiple times)?
    \answerNA{}
	\item Did you include the total amount of compute and the type of resources used (e.g., type of GPUs, internal cluster, or cloud provider)?
    \answerYes{\bf We provide these details in Appendix \ref{implement_det}}
\end{enumerate}

\item If you are using existing assets (e.g., code, data, models) or curating/releasing new assets...
\begin{enumerate}
  \item If your work uses existing assets, did you cite the creators?
    \answerYes{\bf We cite the existing assets in Section \ref{dataset_det} and Appendix \ref{implement_det}.} 
  \item Did you mention the license of the assets?
    \answerYes{\bf We mention these details in Appendix \ref{implement_det}}
  \item Did you include any new assets either in the supplemental material or as a URL?
    \answerYes{\bf We include the GitHub link to our data and code in the Supplementary material}
  \item Did you discuss whether and how consent was obtained from people whose data you're using/curating?
    \answerYes{\bf We mention in Appendix \ref{implement_det}}
  \item Did you discuss whether the data you are using/curating contains personally identifiable information or offensive content?
    \answerYes{\bf We mention in Appendix \ref{implement_det}}
\end{enumerate}

\item If you used crowdsourcing or conducted research with human subjects...
\begin{enumerate}
  \item Did you include the full text of instructions given to participants and screenshots, if applicable?
    \answerYes{\bf See Figure \ref{fig:human-eval-ins}}
  \item Did you describe any potential participant risks, with links to Institutional Review Board (IRB) approvals, if applicable?
    \answerNA{}
  \item Did you include the estimated hourly wage paid to participants and the total amount spent on participant compensation?
    \answerYes{\bf we did not hire any participants for the study and the evaluation is conducted in the research group.}
\end{enumerate}

\end{enumerate}

%%%%%%%%%%%%%%%%%%%%%%%%%%%%%%%%%%%%%%%%%%%%%%%%%%%%%%%%%%%%

%------------------------------------------------------------
\newpage
\section{Related Work}
\label{related_works}
%------------------------------------------------------------
% \newpage
VLMs and ULMs have achieved impressive results on a range of vision and language downstream tasks. These state-of-the-art VLMs and ULMs serve as foundation models for both multimodal applications, like image captioning \cite{li2023blip2}, semantic segmentation of images \cite{Ding0XD22, LiangWDLZ0ZVM23}, text-to-image generation \cite{dalle, dalle-2, imagen}, and unimodal applications, like clustering \cite{wang2023improving, wang2022text}, reranking \cite{bge_embedding}, and retrieval \cite{li2023angle}. 
Their emergence as foundation models has motivated recent research to evaluate the strengths and weaknesses of these models. We summarize the findings from common benchmarks of VLMs and ULMs below. 

\textbf{Findings from the Existing Benchmark for VLMs: }
 \citet{thrush2022winoground} evaluate VLMs through an image-text retrieval task and find that SOTA VLMs struggle to distinguish between texts containing the same words but ordered differently. Similarly, \citet{yuksekgonul2023and} evaluate VLMs in terms of their abilities to form object-attribute associations and highlight shortcomings of VLMs. Other studies with similar conclusions include \cite{zhao2022vl}, \cite{ray2023cola} and \cite{wang2023can}. Recent works have introduced benchmarks to evaluate different abilities of VLMs such as counter-intuitive reasoning \cite{rome_dataset}, visual question answering \cite{xu2023lvlm}, conceptual understanding \cite{schiappa2023probing}, visio-linguistic reasoning \cite{chow2023travlr}, visual-spatial reasoning \cite{liu2023visual} and compositionality \cite{thrush2022winoground}. \citet{kamath2023text} demonstrate challenges in decoding salient aspects of input text encoded with CLIP and draw connections to the lack of compositionality in CLIP text embeddings.

 The task of evaluating compositionality in VLMs is the nearest neighbour to our work. Several datasets have been introduced to evaluate the compositionality of VLMs \cite{liu2023visual, hsieh2023sugarcrepe, thrush2022winoground, zhao2022vl, yuksekgonul2023and, ma2023crepe, ray2023cola, wang2023can, sahin2024enhancing}. Most of the existing compositionality benchmarks formulate the evaluation task as image-text retrieval. Winoground \cite{thrush2022winoground} is one of the earliest benchmarks to report the lack of compositional understanding in VLMs. Latest benchmarks encompassing different aspects of compositionality include VL-CheckList~\cite{zhao2022vl}, CREPE~\cite{ma2023crepe}, Cola~\cite{ray2023cola}, and ARO~\cite{yuksekgonul2023and}. Some benchmarks like Winoground have challenges beyond compositionality that include additional visual and textual reasoning \cite{DiwanBCHM22}. 

\textbf{Findings from the Existing Benchmarks for ULMs:} In the context of ULM text encoders, paraphrasing is the closest to our paper. Paraphrasing is a well-studied problem in NLP. Several previous studies analyzed the ability of the language models to recognize paraphrasing in text. The Microsoft Research Paraphrase Corpus (MRPC) \cite{dolan2005automatically} and Quora Question Pairs (QQP) \cite{qqp_link} are popular paraphrasing datasets (text-only without images) that are part of the GLUE (General Language Understanding Evaluation) \cite{WangSMHLB19} benchmark. The Semantic Textual Similarity (STS) benchmark \cite{sts17t1} build from the STS shared tasks \cite{sts12t6,sts13,sts14t10,sts15t2,sts16t1} have pairs of text snippets with scores indicating the degree of semantic equivalence between them. 

\textbf{Shortcomings of existing Benchmarks:}
\citet{alper2023bert} find that the CLIP text encoder outperforms the ULMs in tasks that require implicit visual reasoning, while \citet{ChenCDWW23} find that ULMs perform better in terms of general language understanding. 
Most VLM benchmarks are generated using rule-based algorithms \cite{ma2023crepe, yuksekgonul2023and} and consist of only a pair of sentences (either semantically similar or dissimilar sentences). These similar pairs might not have strong semantic similarities, and the dissimilar pairs can have significant lexical differences, which does not represent a strict setting of evaluation.  Moreover,  we must finetune or linearly probe \cite{liu2023visual} these models to evaluate LLM encoders using these datasets, which can require significant resources. None of the existing benchmarks systematically evaluates the resilience of model embeddings in the presence of lexical distractors~\cite{iimura2018distractor, taladngoen2022assumptions}, i.e., lexically similar but semantically different negative inputs.

%------------------------------------------------------------
\newpage
\section{\dset Benchmark Generation}
\label{appendix: benchmark generation}
\subsection{Dataset Guidelines}
\label{guidelines}
The main guidelines followed to create the benchmarks are:
\begin{itemize}
\item The lexical changes allowed for creation of the three captions include, replacing words with synonyms and antonyms, reordering the words, etc. These lexical changes do not include adding more details about the image in the caption.
\item Due to the lexical alterations, the three captions should not consist of any nonsensical and non-fluent errors.
\item The three captions should be generated such that they do not need any visual, logical or commonsense reasoning to distinguish the semantically similar captions (P$_1$ and P$_2$) from the semantically different caption (N) i.e., given only three captions without image, one should be able to distinguish P$_1$, P$_2$ from N. 

\end{itemize}

\subsection{Prompt for \dset}
\label{appendix:generic-data-prompt}

Figure \ref{fig:prompting} shows the prompts used to condition the generation of P$_2$ using LLM, given P$_1$. Here, we use instruct fine-tuned Mistral 7B \cite{jiang2023mistral} model to generate P$_2$.

\begin{figure}[!htb]
\
    \centering
    \begin{tcolorbox}[
        colback=gray!5, 
        colframe=gray,
        arc=5mm,
        boxrule=1pt,
    ]
    \textbf{Rules Instruction:} Given an input sentence describing an image caption, follow these steps: \\
    \begin{compactenum}
        \item Rephrase each provided sentence, focusing on preserving the original spatial relationship.
        \item Pay careful attention to the positioning of objects or entities in relation to one another.
        \item Ensure that the meaning remains consistent and that both the original and paraphrased sentences maintain logical coherence and grammatical correctness.
    \end{compactenum}
    \vspace{0.2cm}
    \textbf{Demonstration:} For example, \\
    
    \textbf{\graybox{Input:}} Cat is under the table. \\
    \textbf{\graybox{Paraphrase Idea:}} Rephrase the sentence to convey that the table is positioned above the cat.\\
     \textbf{\graybox{Paraphrased:}} The table is above the cat.\\

    \textbf{\graybox{Input:}}The plane flies below the bright white clouds.\\
    \textbf{\graybox{Paraphrase Idea:}} Ensure the spatial context is maintained by stating that the bright white clouds are situated above the flying plane.\\
     \textbf{\graybox{Paraphrased:}} The plane flies below the bright white clouds.s
    
    \textbf{\graybox{Input:}} The third balcony is below the fourth balcony from the bottom.\\
    \textbf{\graybox{Paraphrase Idea:}} Emphasize the consistent spatial arrangement while indicating that the fourth balcony is positioned above the third balcony from the bottom.\\
     \textbf{\graybox{Paraphrased:}} The fourth balcony is above the third balcony from the bottom.\\

    \textit{Remember to keep the meaning intact, and both the original and paraphrased sentences should be logically coherent and grammatically correct.} \\

    \textbf{\graybox{Input:}} \textit{[Original caption goes here]} \\
    \textbf{\graybox{Paraphrase Idea:}} Focus on replicating the spatial arrangement while maintaining the original meaning of the sentence, correct grammar, same meaning. \\
    \textbf{\graybox{Paraphrased:}} \textit{[Your paraphrased sentence goes here]}
    \end{tcolorbox}
    % \vspace{-8pt}
    \caption{Rules and demonstration sub-prompts used to condition the generator LLM.}
    \label{fig:prompting}
\end{figure}

\newpage
\subsection{Validation prompt for \dset}
\label{app:val_prompt}
Figure \ref{fig:LLM-evaluation} shows the comprehensive prompt used to validate the samples generated by priming the LLM. The outputs obtained from this prompt are further validated by a human expert. This reduces the manual effort required to create the \dataset.
\begin{figure}[ht]
    \centering
    \vspace{0.5cm}
    \begin{tcolorbox}[
        colback=gray!5, 
        colframe=gray,
        arc=5mm,
        boxrule=1pt,
    ]
    \textbf{Instruction:} Given a pair of captions, you must check if the second caption is semantically consistent with the first caption.
    If the second caption is consistent, output the second caption as is; otherwise, rephrase the second caption to be consistent with the first sentence. We are interested in preserving the spatial consistency and spatial relationship of the objects with each other.\\
    
    \textbf{Demonstrations:} examples, \\
    
    \textbf{\graybox{Caption 1:}}  A guy holding a skateboard is speaking into a microphone.\\
    \textbf{\graybox{Caption 2:}}  The guy holding the microphone is speaking into the skateboard.\\
    \textbf{\graybox{isConsistent}}: No, you cannot speak into a skateboard.\\
    \textbf{\graybox{newCaption}}: The guy is speaking into the microphone while holding a skateboard. \\
    
    \textbf{\graybox{Caption 1:}} A family are playing frisbee on the beach.\\
    \textbf{\graybox{Caption 2:}} The frisbee is being played on the beach by a family.\\
    \textbf{\graybox{isConsistent}}: Yes, caption 2 is consistent as it is the same caption written in passive voice. new caption is the same as caption 2.\\
    \textbf{\graybox{newCaption}}: A family are playing frisbee on the beach.\\
    
     \textbf{\graybox{caption 1:}}A stop sign vandalized with an "eating animals" sticker below the word "stop."\\
    \textbf{\graybox{caption 2:}} The stop sign is below an "eating animals" sticker.\\
    \textbf{\graybox{isConsistent}}: The stop cannot be below and above the sticker at the same time.\\
   \textbf{\graybox{newCaption}}: The word "stop" sign is above an "eating animals" sticker.\\
    
     \textbf{\graybox{caption 1:}}There is a phone on top of a calculator.\\
    \textbf{\graybox{caption 2:}} A calculator lies beneath the phone.\\
     \textbf{\graybox{isConsistent}}:Yes, the sentences are semantically equivalent. new caption is same as caption 2.\\
    \textbf{\graybox{newCaption}}: A calculator lies beneath the phone.\\
    
    Now the same for the below caption only.\\
    \textbf{\graybox{caption 1:}} \textit{[Original caption goes here]} \\
   \textbf{\graybox{caption 2:}} \textit{[Generated caption goes here]} \\
    \textbf{\graybox{isConsistent}}: \textit{[Output Here]}
    
    \end{tcolorbox}
    % \vspace{-8pt}
    \caption{Validation Meta-prompt used to validate the consistency of the generated caption and original caption. We use the \textit{isConsistent} output to signal the regeneration of a semantically inconsistent caption.}
    \label{fig:LLM-evaluation}
\end{figure}

\newpage
\subsection{Incorrect generation artifacts by LLMs}
\label{err_analysis_llm}

\begin{figure}[!htb]
    \centering
    \includegraphics[width=\linewidth]{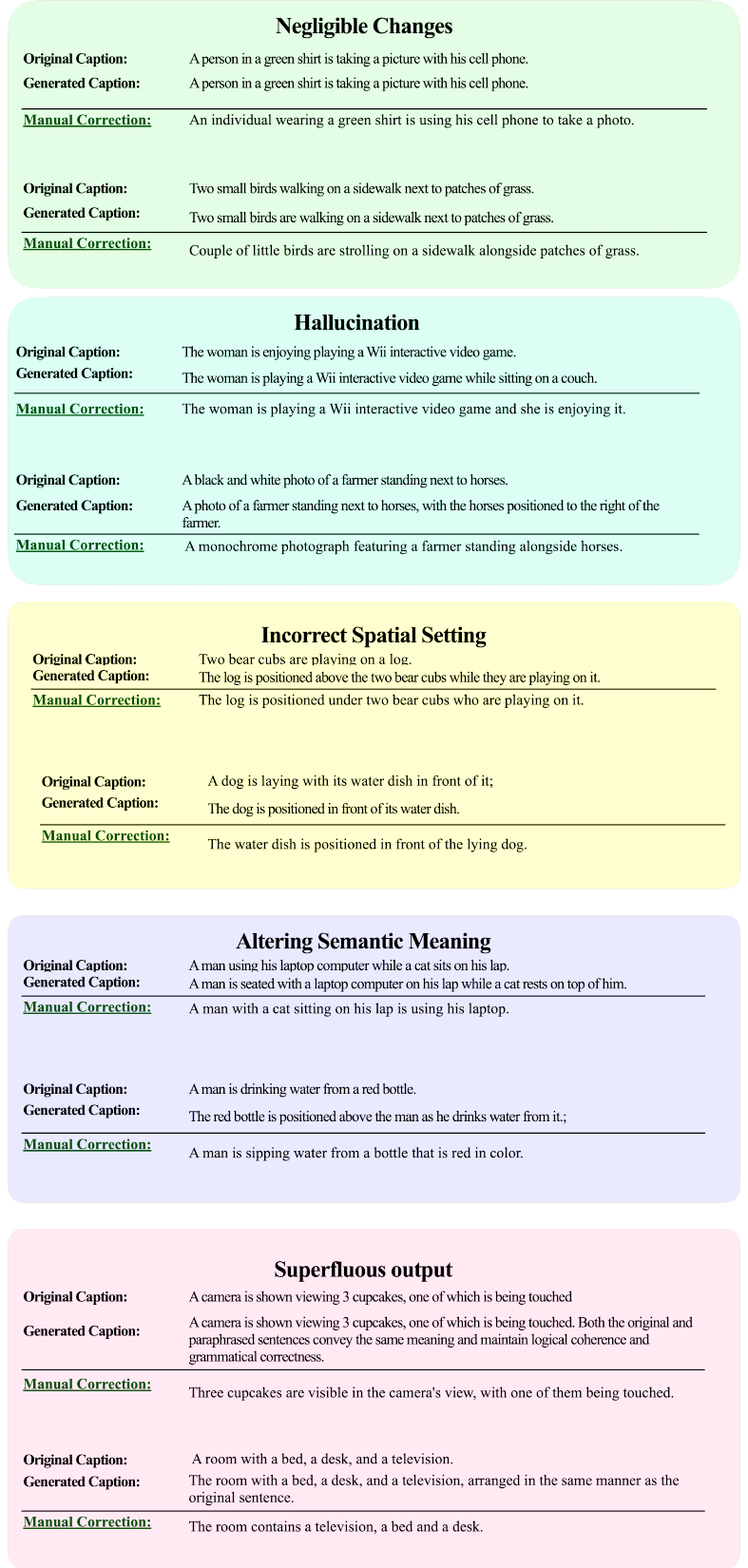}
    \caption{Examples  of \textbf{common errors in the LLM generated positive sentences (P$_2$)}. We provide five of the most common types of errors. Manual correction refers to the corrected sentences after the human validation step. Expert human annotators carefully checked each output of the LLM and corrected the erroneous sentences to maintain grammatical and semantic equivalence with the original positive caption (P$_1$)}
    \label{fig:errors_llm}
\end{figure}

As shown in Figure \ref{fig:errors_llm}, We observe various generation artifacts in the positive sentences (P$_2$) generated by LLMs that were corrected during the human validation stage.

\newpage
%------------------------------------------------------------
\newpage
\section{Additional Analysis of VLMs}
%%%%%%%%%%%%%  Table with details of the VLM models  %%%%
\begin{table}[!htb]
\vspace{0.5cm}
\caption{Details of the VLMs evaluated on the \dataset. Pretraining Data type: R, N and S refer to Real, Noisy and Synthetic data types, respectively. Pretraining Objectives -- ITC: image-text contrastive; ITM: image-text matching; MLM: masked language modeling; MMM: masked multimodal modeling; MIM: masked image modeling; IC: image captioning; IS: image segmentation using KL divergence; ITA: image-text alignment; CCL: Cycle-consistency loss; finetuning objectives -- ITR: image-text retrieval; NL: Negative loss for text; SG: scene graph loss; PT, FT refer to pretraining and finetuning, respectively. CLIP refers to CLIP-ViT-B-32 model.}
\label{tab:vlm_details}
\vspace{0.3cm}
\footnotesize
\resizebox{1.04\textwidth}{!}{
\begin{tabular}{l|lllllc}
\toprule
\multirow{2}{*}{Model} & \#Total & Embedding & Pretraining & Pretraining & Pretraining & \\
& Parameters & Dimension & Data size & Data Type & Objectives & Finetuned \\
\midrule 
CLIP  \citeyear{radford2021learning} & 151M & 512 & 400M & R & ITC & \xmark \\
RoBERTa-ViT-B-32 \citeyear{schuhmann2022laion} & 212M & 512 & 2B & R & ITC &  \xmark \\
ALIGN \citeyear{jia2021scaling} & 490M & 640 & 1.8B & R+N & ITC & \xmark \\
ALIP \citeyear{yang2023alip}  & 151M & 512 & 15M & R+S & ITC & \xmark \\
\midrule
\multirow{2}{*}{FLAVA \citeyear{singh2022flava}}  & \multirow{2}{*}{358M} & \multirow{2}{*}{768} & \multirow{2}{*}{70M} & \multirow{2}{*}{R} & ITC, ITM, MLM & \multirow{2}{*}{\xmark} \\
 & & & & & MMM, MIM & \\
 \hdashline
ALBEF \citeyear{li2021albef} & 210M & 256 & 14M & R+N & ITC, ITM, MLM & \xmark  \\
BLIP \citeyear{li2022blip}  & 225M & 512 & 129M & R+S & ITC, ITM, IC & \xmark \\
BLIP2 \citeyear{li2023blip2}  & 1173M & 256 & 129M & R+S & ITC, ITM, IC & \xmark  \\
\midrule
ViLT \citeyear{kim2021vilt}  & 111M & 768 & 10M & R & ITM, MLM & \xmark  \\
AltCLIP \citeyear{ChenLZYW23}  & 864M & 768 & 42M & R & ITC & \xmark  \\
\midrule
SegCLIP \citeyear{luo2023segclip} & 151M & 512 & 400M+4M & R & ITC, MIM, IS & \xmark \\
XVLM-4M \citeyear{zeng2022multi} & 216M & 256 & 4M & R & ITC, ITM, MLM, ITA & \xmark \\
XVLM-16M \citeyear{zeng2022multi} & 216M & 256 & 16M & R & ITC, ITM, MLM, ITA & \xmark \\
\midrule
% \multirow{2}{*}{BLIP-ITM-COCO \citeyear{li2022blip}}  & \multirow{2}{*}{223M} & \multirow{2}{*}{512} & PT: 129M & R+S & ITC, ITM, IC & \multirow{2}{*}{\cmark} \\
% &&& FT: 110K & R & FT: ITM & \\
% \hdashline
\multirow{2}{*}{ViLT-ITR-COCO \citeyear{kim2021vilt}}  & \multirow{2}{*}{111M} & \multirow{2}{*}{768} & PT: 10M & \multirow{2}{*}{R} & ITM, MLM & \multirow{2}{*}{\cmark} \\
&&& FT: 110K &  & FT: ITR & \\
\hdashline
\multirow{2}{*}{XVLM-16M-COCO \citeyear{zeng2022multi}} & \multirow{2}{*}{216M} & \multirow{2}{*}{256} & PT:16M & \multirow{2}{*}{R} & ITC, ITM, MLM, ITA & \multirow{2}{*}{\cmark}  \\
&&& FT: 110K &  & FT: ITR & \\
\hdashline
\multirow{2}{*}{XVLM-16M-Flickr \citeyear{zeng2022multi}} & \multirow{2}{*}{216M} & \multirow{2}{*}{256} & PT: 16M & \multirow{2}{*}{R} & TC, ITM, MLM, ITA & \multirow{2}{*}{\cmark}  \\
&&& FT: 30K &  & FT: ITR & \\
\midrule
\multirow{2}{*}{NegCLIP \citeyear{yuksekgonul2023and}}  & \multirow{2}{*}{151M} & 512 & PT: 400M & \multirow{2}{*}{R} & ITC & \multirow{2}{*}{\cmark} \\
& & & FT:110K &  & FT: ITM & \\
\hdashline
\multirow{2}{*}{CLIP-SVLC \citeyear{doveh2023teaching}}  & \multirow{2}{*}{151M} & \multirow{2}{*}{512} & PT:400M & \multirow{2}{*}{R} & ITC & \multirow{2}{*}{\cmark}  \\
& & & FT:400M &  & FT: ITC, NL & \\
\hdashline
\multirow{2}{*}{BLIP-SGVL \citeyear{HerzigMKAFDG23}}  & \multirow{2}{*}{696M} & \multirow{2}{*}{768} & PT: 129M & \multirow{2}{*}{R} & ITC, ITM, IC & \multirow{2}{*}{\cmark} \\
& & & FT:4M &  & FT: ITC, SG & \\
\hdashline
CyCLIP \citeyear{goel2022cyclip}  & 102M & 1024 & PT: 102M & R & ITC, CCL &  \xmark \\
\bottomrule
% \vspace{1cm}
\end{tabular}}
\vspace{1cm}
\end{table}

\subsection{Specification of Evaluated VLMs}
\label{appendix:vlm_eval}

We comprehensively evaluate a wide array of VLMs on \dset including (Table \ref{tab:vlm_details} provides details about different VLMs.):

\begin{itemize}
\item Models trained with a contrastive learning objective such as CLIP~\cite{radford2021learning}, RoBERTa-ViT-B-32~\cite{schuhmann2022laion}, ALIGN~\cite{jia2021scaling} and ALIP~\cite{yang2023alip}. ALIGN and ALIP utilize noisy and synthetic captions, respectively. 

\item Models trained by combining multiple objective functions, such as FLAVA~\cite{singh2022flava}: pretrained by combining contrastive, Image-text matching (ITM), masked image modeling (MIM) and masked language modeling (MLM) objectives;  ALBEF~\cite{li2021albef}: which combines ITM and MLM; BLIP~\cite{li2022blip} and BLIP-2~\cite{li2023blip2}: which combine contrastive, ITM and image captioning objectives.

\item Models with a unified encoder for text and images, such as ViLT~\cite{kim2021vilt}, and multi-lingual distilled models like AltCLIP~\cite{ChenLZYW23} 

\item Models that align text with corresponding visual concepts in the image, such as SegCLIP~\cite{luo2023segclip}, and XVLM~\cite{zeng2022multi} - with two variants, XVLM-4M and XVLM-16M.

\item We also evaluate several models that have been finetuned on downstream tasks of image-text retrieval, such as ViLT-ITR-COCO \cite{kim2021vilt} and XVLM-16M-ITR-COCO \cite{zeng2022multi}. Specifically, ViLT, and XVLM-16M models were trained for the ITM task using the COCO dataset. Additionally, XVLM-16M-ITR-Flickr \cite{zeng2022multi} denotes XVLM-16M models trained for the ITM task using the Flickr dataset. 

\item Moreover, we evaluate recent methods proposed to improve the compositionality of VLMs, including NegCLIP \cite{yuksekgonul2023and}, SVLC \cite{doveh2023teaching}, CyCLIP \cite{goel2022cyclip}, and BLIP-SGVL \cite{HerzigMKAFDG23}. 
\end{itemize}

\vspace{0.5cm}

\subsection{Evaluation of variants of CLIP.}
\label{app:clip_var}
We evaluated variants of CLIP~\cite{radford2021learning} that are different in pre-training data size, model architecture and model size as listed below (see Table \ref{tab:clip_variants} for more details).
\begin{itemize}
\item CLIP~\cite{radford2021learning} variants trained on the WebImageText dataset, which comprises 400 million image-text pairs. These models encompass CNN-based architectures, such as RN50, RN101, RN$50\times4$, RN$50\times16$, and RN$50\times64$, as well as transformer-based models like ViT-B/32 and ViT-L/14. 
\item CLIP-based models introduced by \cite{schuhmann2022laion} pre-trained on extensive paired image-text datasets. \citet{schuhmann2022laion} provided diverse CLIP variants, namely RoBERTa-ViT-B/32, ViT-H/14, ViT-g/14, xlm-roberta-base-ViT-B/32, and xlm-roberta-large-ViT-H/14, trained on a large image-text dataset called "LAION-5B," which consists of 5 billion image-text pairs. 
\item \citet{gadre2023datacomp} released two CLIP variants, namely Large:ViT-B/16 and xlarge:ViT-L/14, trained on the DataComp dataset, comprising 13 billion image-text pairs.
\end{itemize}

Performance of various CLIP variants on \dataset is provided in Table \ref{clip_var_sc_pp_comp}.

% \clearpage
\begin{table}[!htb]
\caption{\small Details of different variants of CLIP that are evaluated on the \dataset. Data, Model and Emb. refer to the pre-training dataset size and total number of parameters in the model (in Millions) and embedding dimension, respectively.}
\label{tab:clip_variants}
\vspace{0.5cm}
\footnotesize
\centering
\begin{tabular}{llrrr}
\toprule
& Pre-training& Pre-training & \# Params & Embed.\\
Model & Dataset &Data size & Model & Dimen.\\
\midrule 
RN50 & WebImageText & 400M & 102M & 1024 \\
RN101 & WebImageText & 400M & 120M & 512 \\
CLIP & WebImageText & 400M & 151M & 512  \\
RN50$\times$4 & WebImageText & 400M & 178M & 640 \\
RN50$\times$16 & WebImageText & 400M & 291M & 768 \\
CLIP-ViT-L/14 & WebImageText & 400M & 428M & 768 \\
RN50$\times$64 & WebImageText & 400M & 623M & 1024 \\
\midrule
RoBERTa-ViT-B/32 & LAION & 2B & 212M & 512 \\
ViT-H/14 & LAION & 2B & 986M & 1024 \\
ViT-g/14 & LAION & 2B & 1367M & 1024 \\
ViT-bigG/14 & LAION & 2B & 2540M & 1280 \\
xlm-roberta-base-ViT-B/32 & LAION & 5B & 366M & 512 \\
xlm-roberta-large-ViT-H/14 & LAION & 5B & 1193M & 1024 \\
\midrule
large:ViT-B/16 & DataComp & 1B & 150M & 512 \\
xlarge:ViT-L/14 & DataComp & 13B & 428M & 768 \\
\bottomrule
\end{tabular}
\vspace{1cm}
\end{table}

\begin{table}[!tb]
\caption{\small Comparison of the performance of different variants of CLIP on \dset. Performance reported in terms of Accuracy (\%). Overall best values are in bold, and group-level best values are underlined.}
\label{clip_var_sc_pp_comp}
\vspace{0.5cm}
\footnotesize
\centering
\resizebox{1.01\linewidth}{!}{
\begin{tabular}{lcccccccccc}
\toprule
Model & \multicolumn{2}{c}{Swap Object} & \multicolumn{2}{c}{Swap Attribute} & \multicolumn{2}{c}{Replace Object} & \multicolumn{2}{c}{Replace Attribute} & \multicolumn{2}{c}{Replace Relation} \\
\cmidrule(lr){2-3} \cmidrule(lr){4-5} \cmidrule(lr){6-7} \cmidrule(lr){8-9} \cmidrule(lr){10-11}
 & \multicolumn{1}{c}{\ITT} & \multicolumn{1}{c}{\TOT} & \multicolumn{1}{c}{\ITT} & \multicolumn{1}{c}{\TOT} & \multicolumn{1}{c}{\ITT} & \multicolumn{1}{c}{\TOT} & \multicolumn{1}{c}{\ITT} & \multicolumn{1}{c}{\TOT} & \multicolumn{1}{c}{\ITT} & \multicolumn{1}{c}{\TOT} \\
 \midrule
RN50 \cite{radford2021learning} & 46.05 & 17.98 & \underline{49.01} & 31.35 & 87.41 & 82.45 & 67.39 & 57.36 & 56.12 & 39.04 \\
RN101 \cite{radford2021learning} & 42.10 & 18.86 & 48.10 & 29.68 & 88.20 & 83.17 & \underline{67.89} & 55.21 & 53.20 & \underline{39.76} \\
CLIP \cite{radford2021learning} & 45.18 & 19.74 & 45.21 & \underline{33.03} & 86.80 & \underline{83.72} & 65.61 & \underline{59.14} & \underline{56.26} & 38.62 \\
RN50$\times$4 \cite{radford2021learning} & \textbf{46.93} & \underline{21.49} & 46.42 & 30.59 & 87.77 & 80.87 & 67.51 & 53.93 & 53.91 & 38.55 \\
RN50$\times$16 \cite{radford2021learning} & 39.04 & 17.55 & 46.42 & 30.29 & 89.10 & 76.57 & 65.74 & 49.87 & 53.41 & 38.19 \\
CLIP-ViT-L/14 \cite{radford2021learning} & 43.86 & 17.11 & 45.36 & 28.92 & 90.68 & 80.69 & 67.39 & 55.96 & 54.05 & 39.26 \\
RN50$\times$64 \cite{radford2021learning} & 44.74 & 16.67 & 45.36 & 31.51 & \underline{90.79} & 73.31 & 64.47 & 48.61 & 54.27 & 38.12 \\
\midrule
RoBERTa-ViT-B/32 \cite{schuhmann2022laion} & 44.3 & 29.39 & \underline{56.32} & 52.66 & 89.04 & 94.55 & 74.49 & 80.46 & 59.39 & 57.75 \\
ViT-H/14 \cite{schuhmann2022laion} & 43.42 & 27.63 & 54.19 & 50.69 & 93.71 & 90.43 & \underline{71.06} & 73.98 & 56.62 & \underline{51.92} \\
ViT-g/14 \cite{schuhmann2022laion} & 44.3 & 26.32 & 52.06 & \underline{43.99} & 93.1 & \underline{91.83} & 71.19 & \underline{73.73} & \underline{57.54} & 52.56 \\
ViT-bigG/14 \cite{schuhmann2022laion} & \underline{45.61} & \underline{29.82} & \textbf{57.38} & 52.05 & \textbf{94.13} & 90.44 & \textbf{76.40} & 72.84 & \textbf{59.45} & 53.49 \\
xlm-RoBERTa-base-ViT-B/32 \cite{schuhmann2022laion} & 42.55 & \textbf{30.26} & 55.25 & \textbf{55.56} & 89.41 & \textbf{95.34} & 72.97 & \textbf{80.96} & 55.48 & 57.82 \\
xlm-RoBERTa-large-ViT-H/14 \cite{schuhmann2022laion} & 42.1 & 29.83 & 54.19 & 52.36 & 93.94 & 94.25 & 75.38 & 79.19 & 58.75 & \textbf{60.69} \\
\midrule
large:ViT-B/16 \cite{gadre2023datacomp} & 35.96 & 17.98 & 39.58 & 30.44 & 87.83 & 90.5 & 67.89 & 70.81 & 50.36 & 39.04 \\
xlarge:ViT-L/14 \cite{gadre2023datacomp} & \underline{42.55} & \underline{26.75} & \underline{46.58} & \underline{39.12} & \underline{91.59} & \underline{91.89} & \underline{72.59} & \underline{71.45} & \underline{55.83} & \underline{49.79} \\
\bottomrule
\end{tabular}}
\vspace{0.5cm}
\end{table}

%------------------------------------------------------------
\newpage
\section{Additional Analysis of ULMs}
\label{appendix:addtional-analysis-ULMs}
%------------------------------------------------------------
% 4 3 8
% 9 5 1
% 2 7 6
Unimodal Language Models (ULMs) can be evaluated for their semantic and lexical sensitivity using the Text-only task (TOT) in \dset. We report the TOT results of ULMs in Table \ref{tab:unimodal-performance}. We cover a comprehensive list of ULMs, varying in parameter counts, optimization objectives, training data sizes, and architectures. We notice that ULMs consistently fail to disassociate semantics from different lexical forms. This is indicated by the large deviation (ranging from 18\% to 35\%) in TOT accuracy across different subsets. Generally, all models perform over 90\% accuracy in identifying the positive captions when the negative caption changes the type of an 'object' in the positive caption (altering the semantics and lowering lexical overlap). On the contrary, if multiple 'objects' are swapped within the same caption (altering the semantics but preserving the lexical overlap), the performance, on average across all ULMs, decreases by $\approx$50\%. This can be attributed to the lack of compositional understanding required to associate semantics with different lexical forms. 

Grouping the models based on model sizes reveals that sensitivity to lexical alteration does not necessarily improve with scale. For example, a small-sized model like BAAI General Embedding (BGE) \cite{bge_embedding} is only 5\% behind very large LLMs like E5-mistral \cite{wang2023improving, wang2022text} with 7 billion parameters in the 'Replace Relation' subset. We experimented with recently proposed text encoder models like INSTRUCTOR \cite{INSTRUCTOR} that aim to improve the generalization of sentence embeddings across various embedding-based tasks. These models allow the addition of special prefix instructions before encoding text to condition for a task. We compared the INSTRUCTOR model with a default prefix and a prefix with instructions to encode the semantics (listed as custom-ins in Table \ref{tab:unimodal-performance}). We notice minimal performance gains with custom instruction, demonstrating that further development is required to separate semantics from lexical composition.

\begin{table*}[!tb]
    \centering
    \caption{\small Comprehensive results of ULMs on \TOT  of \dset. We report the \TOT accuracy  (\%). and group the results row-wise based on the model size as reflected by the parameter count. We include the number of parameters in text encoders relative to BERT-base, i.e., 109.5 Million parameters. Overall, best values are in bold, and group-level best values are underlined. We report the average and standard deviation across different subsets as an additional column.} 
     \label{tab:unimodal-performance}
    \resizebox{\linewidth}{!}{
    \begin{tabular}{@{}lccccccc@{}}
\toprule
Model &  \begin{tabular}[c]{@{}c@{}}\#Params \\ (BERT Scale)\end{tabular} &\begin{tabular}[c]{@{}c@{}}Swap \\ Object\end{tabular} & \begin{tabular}[c]{@{}c@{}}Swap \\ Attribute\end{tabular} & \begin{tabular}[c]{@{}c@{}}Replace \\ Object\end{tabular} & \begin{tabular}[c]{@{}c@{}}Replace \\ Attribute\end{tabular} & \begin{tabular}[c]{@{}c@{}}Replace \\ Relation\end{tabular} & Average \\
            \midrule
            All-MiniLM-L6-v2 \cite{wang2020minilm} &0.21 & 14.29& 22.52& 93.95& 64.97&63.94&51.93$_{33.02}$\\
            BGE-small-en-v1.5 \cite{bge_embedding}&0.3 & 15.51& 24.02& 94.19& \underline{75.00} &\underline{75.53}&\underline{56.85}$_{34.85}$\\
            All-MiniLM-L12-v2 \cite{wang2020minilm} &0.3 & \underline{18.78}& \underline{25.38}& \underline{95.22}& 73.86&70.41&56.73$_{33.11}$\\
            GTE-small \cite{li2023general} &0.3 & 13.88& 22.07& 94.98& 71.95&69.06&54.39$_{34.84}$\\
            \midrule 
            Angle-BERT-base-uncased-nli-en-v1 \cite{li2023angle}  &1  & 25.71& \underline{33.63}& 92.07& 78.43&75.32&\underline{61.03}$_{29.45}$\\
            BGE-base-en-v1.5  \cite{bge_embedding} &1 & 17.14& 25.23& 93.83& \underline{78.55}&\underline{76.10}&58.17$_{34.56}$\\
            Sentence-T5-base \cite{sent-t5} &1.01  & \underline{28.98}& 31.98& 92.37& 75.00&75.32&60.73$_{28.51}$\\
            GTE-base \cite{li2023general} &1 & 17.14& 22.82& \underline{94.31} & 75.00&71.98&56.25$_{34.26}$\\
            \midrule 
            Instructor-large \cite{INSTRUCTOR} &3.07 & 26.53& 28.38& 96.00& 72.34&73.83&59.42$_{30.65}$\\
            Instructor-large(custom-ins)\cite{INSTRUCTOR} &3.07 & 22.86& 27.63& 96.13& 77.41&\underline{77.81}&60.37$_{32.99}$\\
            UAE-Large-v1 \cite{li2023angle} &3.06 & 40.41& 41.44& \textbf{96.85}& 76.14&75.82&66.13$_{24.54}$\\
            GTE-large \cite{li2023general} &3.06 & 26.53& 27.93& 96.31& 76.78&72.83&60.07$_{31.28}$\\
            All-RoBERTa-large-v1 \cite{sbert}  &3.25 & 42.04& 45.20& 94.61& 74.75&74.96&66.31$_{22.26}$\\
            Stsb-RoBERTa-large \cite{sbert} &3.25 & 25.31& 31.98& 94.19& \textbf{89.21}&75.18&63.17$_{32.37}$\\
            Sentence-T5-xl \cite{sent-t5} &11.34 & \textbf{47.35}& \textbf{49.25}& 90.98& 75.38&75.32&\underline{67.66}$_{18.8}$\\
            \midrule 
            Angle-Llama-7b-nli-v2 \cite{li2023angle}&62.28 & \underline{37.96}& \underline{45.80}& 95.22& 84.39&\textbf{81.44}&\textbf{68.96}$_{25.4}$\\
            E5-Mistral-7b-instruct \cite{wang2023improving, wang2022text} &64.95 & 33.47& 37.84& \underline{96.67}& \underline{87.06}&80.51&67.11$_{29.33}$\\
            \bottomrule
        \end{tabular}
        }
     
\end{table*}

\newpage
\section{Human Evaluation Instruction}
\begin{figure}[!htb]
    \centering
    \includegraphics[width=0.9\linewidth]{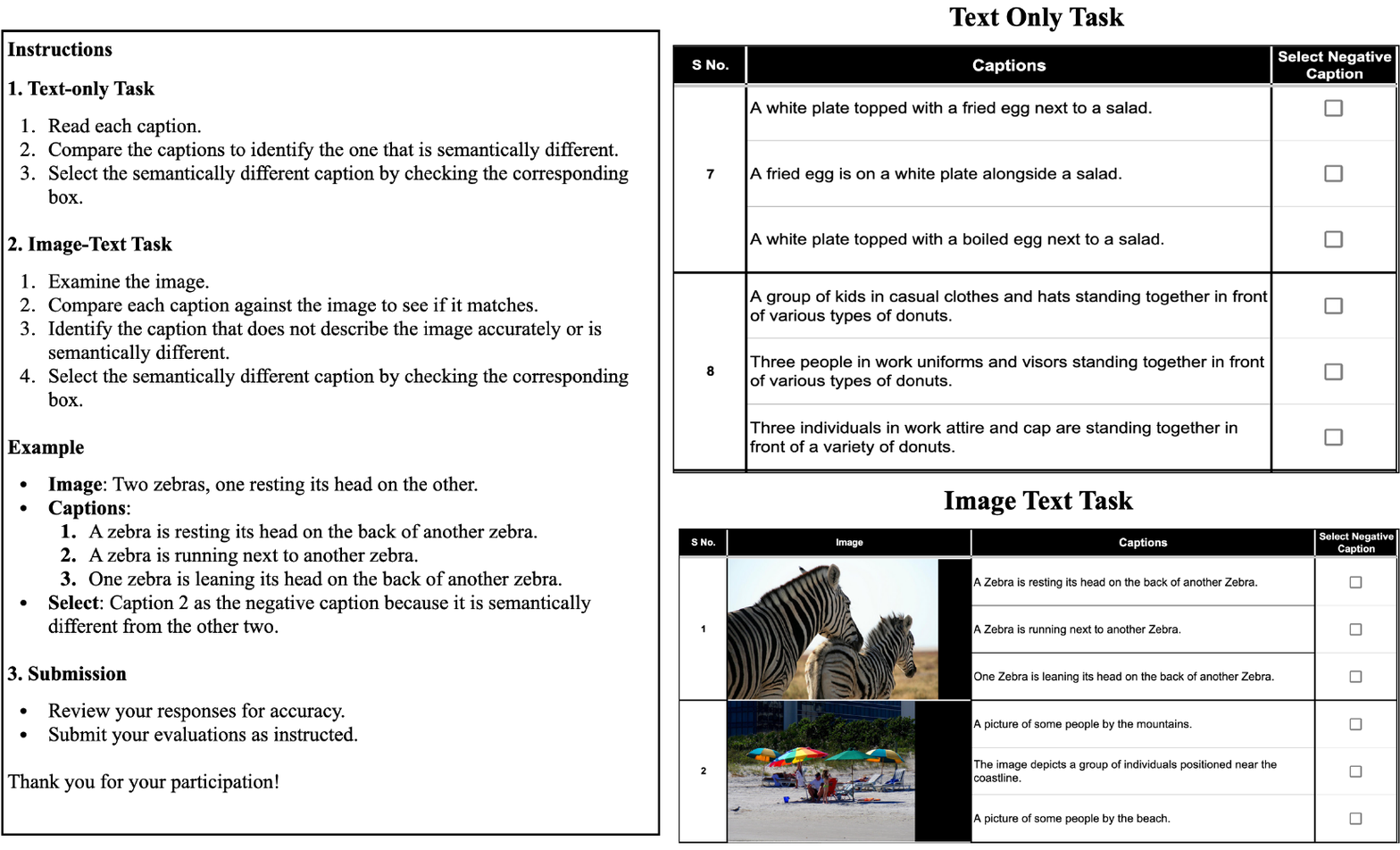}
    \caption{Human evaluation instructions and a screenshot of the interface.}
    \label{fig:human-eval-ins}
\end{figure}

Figure \ref{fig:human-eval-ins} provides the instructions given to the human evaluators along with the screenshot of the interface used for human evaluation. 

%------------------------------------------------------------
% \newpage
\section{Additional Analysis and Results}

\subsection{Prompt-based Evaluation}
We evaluated the latest released GPT-4o ("o" for "omni") model on \dset in a prompt-based format. Given an image, we prompted the GPT-4o API~\footnote{\url{https://platform.openai.com/docs/guides/vision}} using three different prompts.

\begin{itemize}
\item \textbf{Prompt-1:} "Do any of the following captions not match the image? (1) " + <Negative Caption> + "; (2) " + <Positive Caption 1> + "; (3) " + <Positive caption 2> +  "; provide output as (1), (2), (3) or none"
\item \textbf{Prompt-2:} "Do any of these captions fail to correspond with the image? (1) " + <Negative Caption> + "; (2) " + <Positive Caption 1> + "; (3) " + <Positive caption 2> +  "; provide output as (1), (2), (3) or none"
\item \textbf{Prompt-3:} "Do any of these captions fail to correspond with the image? ((1) " + <Negative Caption> + "; (2) " + <Positive Caption 1> + "; (3) " + <Positive caption 2> +  "; provide output as (1), (2) or (3)"
\end{itemize}

In the above prompts, <Negative caption>, <Positive caption 1> and <Positive caption 2> refer to the negative (N) and the two positive captions (P$_1$ and P$_2$) corresponding to the image, respectively. First two prompts (Prompt-1 and Prompt-2) are paraphrases of each other and are 4-class problems (Output to be (1), (2), (3) or none). Whereas Prompt-3 is a 3-class problem i.e., model need to output (1), (2) or (3).

Table \ref{tab:gpt_4o} provides the performance of GPT-4o on \dset for the three different prompts. Differences in performance of GPT-4o are observed between Prompt-1 and Prompt-2, which are paraphrases of one another. This shows the sensitivity of the GPT-4o model to the structure of the prompt. When prompted to choose output from four options ((1), (2), (3), or none), GPT-4o model has difficulty identifying the negative caption (a caption that does not correspond to the image). Conversely, GPT-4o better identified the negative caption when prompted to choose from three options ((1), (2), or (3)). For all the three prompts, GPT-4o was better at identifying the negative prompt when we replaced an object or attribute in the positive caption to form the negative caption. GPT-4o has difficulty in identifying the negative caption when the relation between the objects is replaced or when the objects or attributes are swapped to form the negative caption. For the replace object and replace attribute categories, GPT-4o's performance is close to human performance. However, for swap object, swap attribute, and replace relation cases, GPT-4o's performance lags behind human performance by a large margin.

\begin{table}[htb]
% \vspace{-0.3cm}
\caption{\small Prompt-based evaluation of GPT-4o on \dset. We provide both image and a prompt to the GPT-4o and receive the output from GPT-4o. Based on the response, we compute the performance i.e., it is a hit if the model outputs (1) else a miss. Performance is reported in terms of Accuracy (\%), where accuracy is computed as the ratio of the \#hits/(\#hits + \#misses). }
\label{tab:gpt_4o}
\vspace{0.3cm}
\centering
\footnotesize
\resizebox{1\linewidth}{!}{
\begin{tabular}{lccccc}
\toprule
Model & \multicolumn{1}{c}{Swap Object} & \multicolumn{1}{c}{Swap Attribute} & \multicolumn{1}{c}{Replace Object} & \multicolumn{1}{c}{Replace Attribute} & \multicolumn{1}{c}{Replace Relation} \\
\midrule
 Human & 100.00 & 96.67 & 100.00 & 100.00 & 100.00 \\
 \midrule
 Prompt-1 & 46.93 & 73.36 & 91.64 & 87.94 & 69.06  \\
 Prompt-2 & 48.25 & 75.04 & 90.82 & 84.90 & 71.19  \\
 Prompt-3 & \textbf{67.61} & \textbf{85.82} & \textbf{96.25} & \textbf{93.27} & \textbf{84.13}  \\
\bottomrule
\end{tabular}}
\end{table}

\newpage
\subsection{Qualitative results}
We perform qualitative analysis to inspect examples from \dataset, where the majority of models fail to dissociate semantics from lexical composition. In this section, we highlight the results of our qualitative analysis.

\begin{figure}[!ht]
\vspace{0.5cm}
    \centering
    \includegraphics[width=\textwidth]{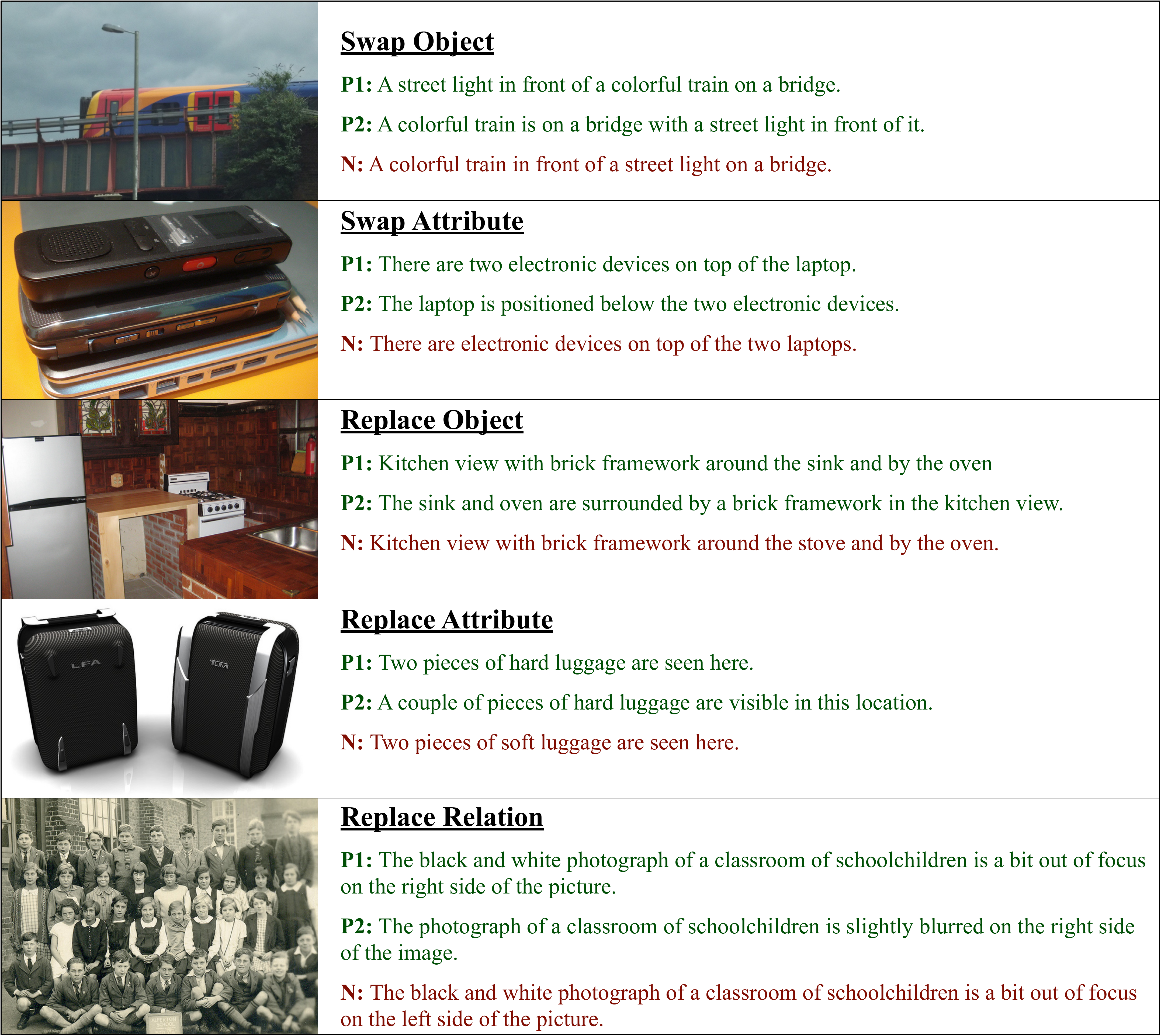}
    \caption{Examples from each subset of \dataset where the majority of VLMs failed both the image-text task (ITT) and the text-only task (TOT).}
    \label{fig:quali-both-fail}
\end{figure}

\subsubsection{Both Image-text task (ITT) and Text-only Task (TOT) fail
}
In this subsection, we provide examples where the majority of vision-language models failed in both the image-text task and the text-only task. These examples demonstrate cases where both the text encoder and vision encoder of VLMs were ineffective in encoding semantics. Figure \ref{fig:quali-both-fail} shows such examples from each subset of \dataset.

\newpage
\subsubsection{Image-text task (ITT) pass, and Text-only task (TOT) fail}
\begin{figure}[!ht]
\vspace{0.9cm}
\hspace{-0.3cm}
    \centering
    \includegraphics[width=1.04\textwidth]{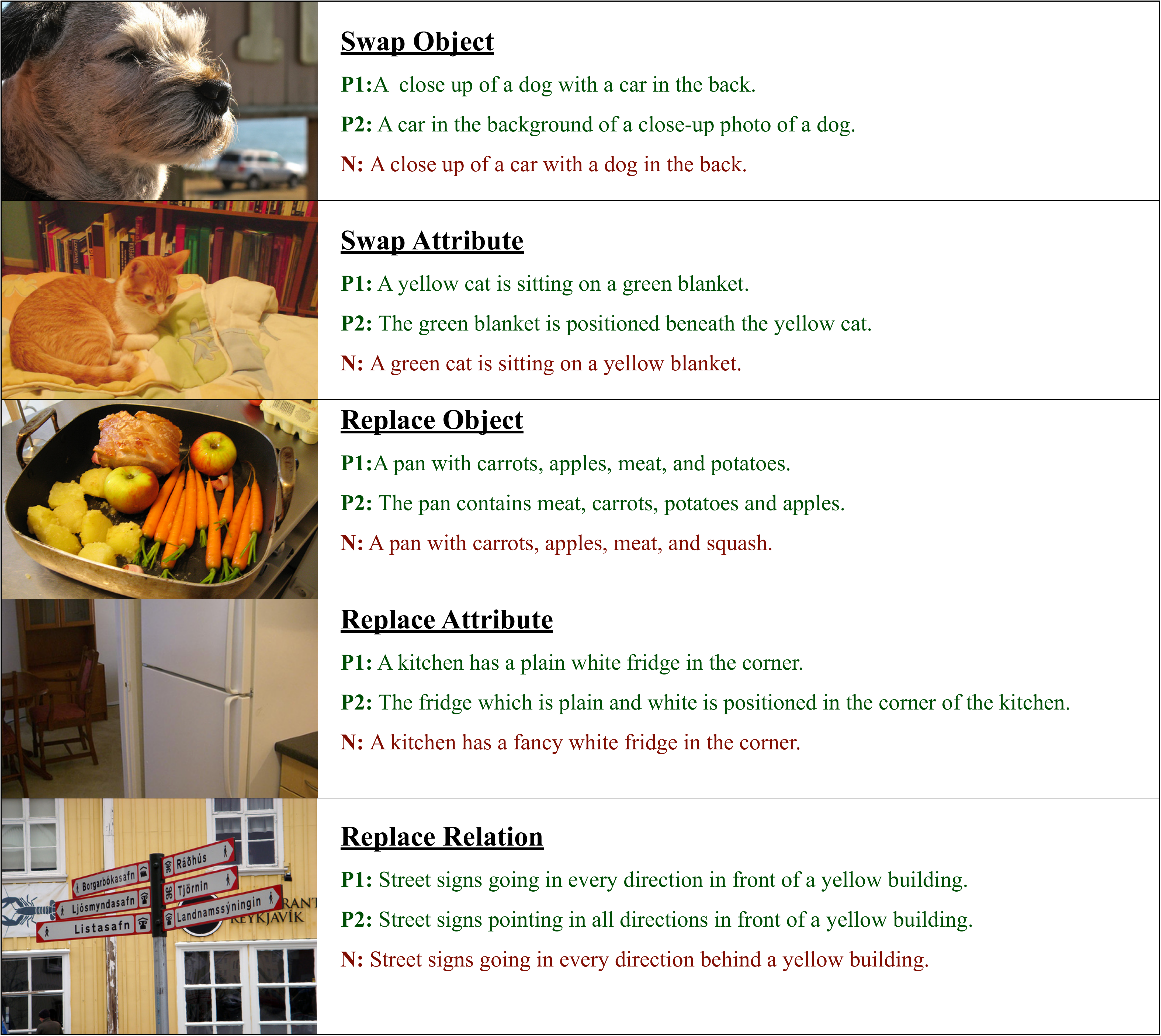}
    \vspace{0.2cm}
    \caption{Examples from each subset of \dataset where the majority of VLMs passed the image-text task (ITT) and failed the text-only task (TOT).}
    \label{fig:quali-itt-pass-tot-fail}
\end{figure}

In this subsection, we provide examples where the majority of vision-language models (VLMs) failed in the text-only task (\TOT) but were successful in the image-text task (\ITT), indicating that the vision encoders were able to represent semantics more effectively than text encoders for these examples. Figure \ref{fig:quali-itt-pass-tot-fail} shows such examples from each subset of \dataset.

\newpage
\subsubsection{Text-only task (TOT) pass and Image-text task fail}
\begin{figure}[!ht]
\vspace{0.9cm}
    \centering
    \hspace{-0.5cm}
    \includegraphics[width=1.06\textwidth]{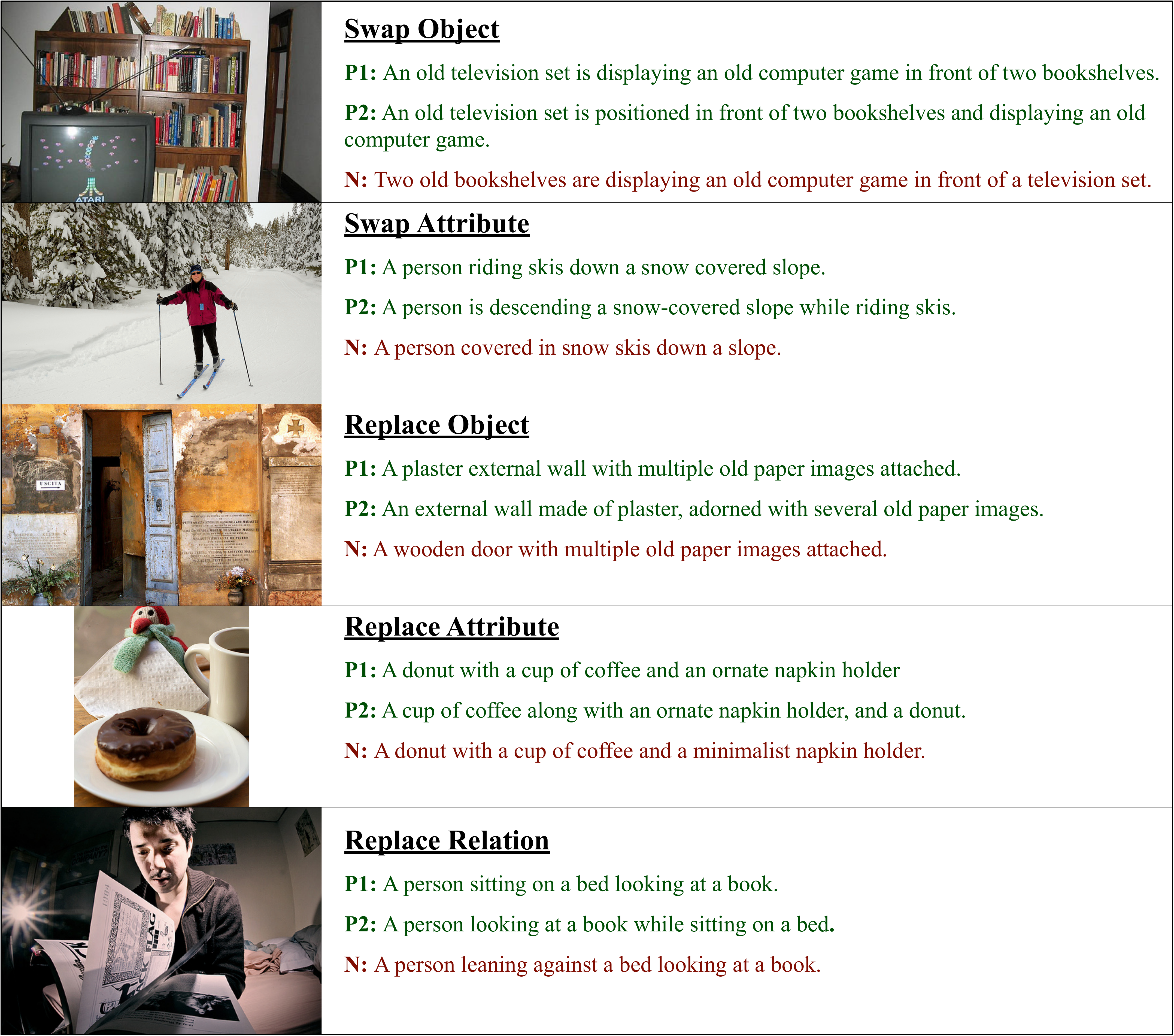}
    \vspace{0.2cm}
    \caption{Examples from each subset of \dataset where the majority of VLMs passed the text-only task (TOT) and failed the image-text task (ITT).}
    \label{fig:quali-itt-fail-tot-pass}
\end{figure}

In this subsection, we provide examples where the majority of vision-language models (VLMs) failed in the image-text task (ITT) but succeeded in the text-only task (TOT). These examples demonstrate cases where the text encoders of VLMs were more effective at encoding semantics than their vision encoders. Figure \ref{fig:quali-itt-fail-tot-pass} presents examples from each subset of \dataset.

%------------------------------------------------------------
\newpage
\section{Implementation Details}
\label{implement_det}

\subsection{Hardware information}
We performed all the experiments in this paper using a single 40G NVIDIA A100 GPU available in the Compute Canada Cluster.

\subsection{Dataset sources}
We obtain all existing datasets from their original sources released by the authors. We refer readers to these sources for the dataset licenses.

\begin{compactitem}
    \item COCO~\cite{lin2014microsoft}: We obtain COCO images from its official project website~\footnote{\url{https://cocodataset.org/}}. We use the images from the validation set~\footnote{\url{http://images.cocodataset.org/zips/val2017.zip}}
    \item \scd~\cite{hsieh2023sugarcrepe}: We obtain \scd captions and hard negatives from its official website~\footnote{\url{https://github.com/RAIVNLab/sugar-crepe}}.
\end{compactitem}

\subsection{Models}
\textbf{Evaluation of VLMs. }
Source and links of the VLMs detailed in Table 8 (in the Appendix of our paper) is provided below.

\begin{itemize}
\item CLIP~\cite{radford2021learning}: 'ViT-B/32' variant of CLIP available at \href{https://huggingface.co/openai/clip-vit-base-patch32}{HuggingFace Link}
\item RoBERTa-ViT-B-32 \cite{schuhmann2022laion}: RoBERTa-ViT-B-32 trained on LAION dataset available at \href{https://huggingface.co/laion/CLIP-ViT-B-32-roberta-base-laion2B-s12B-b32k}{HuggingFace Link}
\item ALIGN  \cite{jia2021scaling}: Model available at \href{https://huggingface.co/kakaobrain/align-base}{HuggingFace Link}
\item ALIP \cite{yang2023alip}: Model available at \href{https://drive.google.com/file/d/1AqSHisCKZOZ16Q3sYguK6zIZIuwwEriE/view?usp=sharing}{Google Drive Link}
\item FLAVA \cite{singh2022flava}: Model available at \href{https://huggingface.co/facebook/flava-full}{HuggingFace Link}
\item ALBEF \cite{li2021albef}: ALBEF base model available in \href{https://github.com/salesforce/LAVIS/tree/main}{LAVIS}
\item BLIP \cite{li2022blip}: BLIP base model available in \href{https://github.com/salesforce/LAVIS/tree/main}{LAVIS}
\item BLIP2 \cite{li2023blip2}: BLIP2 pretrained model available in \href{https://github.com/salesforce/LAVIS/tree/main}{LAVIS}
\item ViLT \cite{kim2021vilt}: Pre-trained ViLT model available at \href{https://huggingface.co/dandelin/vilt-b32-mlm}{HuggingFace Link}
\item SegCLIP \cite{luo2023segclip}: Model available at \href{https://github.com/ArrowLuo/SegCLIP/releases/download/checkpoint_v0/segclip.bin}{Google Drive Link}
\item XVLM-4M \cite{zeng2022multi}: XVLM base model trained using 4 Million samples available at \href{https://drive.google.com/file/d/1B3gzyzuDN1DU0lvt2kDz2nTTwSKWqzV5/view?usp=sharing}{Google Drive Link}
\item XVLM-16M \cite{zeng2022multi}: XVLM base model trained using 16 Million samples available at \href{https://drive.google.com/file/d/1iXgITaSbQ1oGPPvGaV0Hlae4QiJG5gx0/view?usp=sharing}{Google Drive Link}
\item ViLT-ITR-COCO \cite{kim2021vilt}: ViLT model finetuned for image-text retrieval task using MSCOCO dataset. This model is available at \href{https://huggingface.co/dandelin/vilt-b32-finetuned-coco}{HuggingFace Link}
\item XVLM-16M-COCO \cite{zeng2022multi}: XVLM 16 Million model fine-tuned for image-text-retrieval using MSCOCO dataset. This model is available at \href{https://drive.google.com/file/d/1bv6_pZOsXW53EhlwU0ZgSk03uzFI61pN/view?usp=share_link}{Google Drive Link}
\item XVLM-16M-Flickr \cite{zeng2022multi}: XVLM 16 Million model fine-tuned for image-text-retrieval using Flickr dataset. This model is available at \href{https://drive.google.com/file/d/1vhdtH3iFaoZuMqOGm-8YM-diPWVfRJzv/view?usp=share_link}{Google Drive Link}
\item NegCLIP \cite{yuksekgonul2023and}: NegCLIP is trained on top of CLIP and the model link is available in \href{https://github.com/mertyg/vision-language-models-are-bows/tree/main}{Github Link}
\item CLIP-SVLC \cite{doveh2023teaching}: Model is available at \href{https://drive.google.com/file/d/1k-JAVRnyX0UGSY0Ng5EA1vD4GrhbiVZ2/view?usp=share_link}{Google Drive Link}
\item BLIP-SGVL \cite{HerzigMKAFDG23}: Model is available at \href{https://drive.google.com/file/d/13jzpcLgGalO3hkiqVwziNAlCEZD90ENN/view?usp=sharing}{Google Drive Link}
\item CyCLIP \cite{goel2022cyclip}: Model is available at \href{https://drive.google.com/file/d/1nF33F3yjtiWr3bgllBXk5Wf07Uo7Uv9G/view?usp=share_link}{Google Drive Link}
\end{itemize}

% \newpage
\textbf{Variants of CLIP:} All the CLIP models' weights for the different variants of CLIP reported in Table 10 in the appendix are obtained from OpenCLIP~\footnote{\url{https://github.com/mlfoundations/open_clip}} framework~\cite{openclip}.

\textbf{Evaluated ULMs:}
The source and links of the ULMs detailed in Table 11 (in the Appendix of our paper) are provided below.

    \begin{itemize}
\item All-MiniLM-L6-v2 \cite{wang2020minilm}: \href{https://huggingface.co/sentence-transformers/all-MiniLM-L6-v2}{HuggingFace Link}   
\item BGE-small-en-v1.5 \cite{bge_embedding}: \href{https://huggingface.co/BAAI/bge-small-en-v1.5}{HuggingFace Link}                     
\item All-MiniLM-L12-v2 \cite{wang2020minilm}: \href{https://huggingface.co/sentence-transformers/all-MiniLM-L12-v2}{HuggingFace Link}    
\item GTE-small \cite{li2023general}: \href{https://huggingface.co/thenlper/gte-small}{HuggingFace Link}                         
\item Angle-BERT-base-uncased-nli-en-v1 \cite{li2023angle}: \href{https://huggingface.co/SeanLee97/angle-bert-base-uncased-nli-en-v1}{HuggingFace Link} 
\item BGE-base-en-v1.5 \cite{bge_embedding}: \href{https://huggingface.co/BAAI/bge-base-en-v1.5}{HuggingFace Link}                      
\item Sentence-T5-base \cite{sent-t5}: \href{https://huggingface.co/sentence-transformers/sentence-t5-base}{HuggingFace Link}     
\item GTE-base \cite{li2023general}: \href{https://huggingface.co/thenlper/gte-base}{HuggingFace Link}                          
\item Instructor-large \cite{INSTRUCTOR}: \href{https://huggingface.co/hkunlp/instructor-large}{HuggingFace Link} 
\item Instructor-large (custom-ins)\cite{INSTRUCTOR}: \href{https://huggingface.co/hkunlp/instructor-large}{HuggingFace Link}, we use \textit{`Represent the sentence for spatial semantics'} as the custom instruction for Instructor-large (custom-ins) model. 
\item UAE-Large-V1 \cite{li2023angle}: \href{https://huggingface.co/WhereIsAI/UAE-Large-V1}{HuggingFace Link}                     
\item GTE-large \cite{li2023general}: \href{https://huggingface.co/thenlper/gte-large}{HuggingFace Link}                         
\item All-RoBERTa-large-v1 \cite{sbert}: \href{https://huggingface.co/sentence-transformers/all-roberta-large-v1}{HuggingFace Link} 
\item Stsb-RoBERTa-large \cite{sbert}: \href{https://huggingface.co/sentence-transformers/stsb-roberta-large}{HuggingFace Link}   
\item Sentence-T5-xl \cite{sent-t5}: \href{https://huggingface.co/sentence-transformers/sentence-t5-xl}{HuggingFace Link}       
\item Angle-Llama-7b-nli-v2 \cite{li2023angle}: \href{https://huggingface.co/SeanLee97/angle-llama-7b-nli-v2}{HuggingFace Link}            
    \end{itemize}

\subsection{Reproducibility}
\label{reproduce}

We release \dataset and the code to evaluate models on Github~\footnote{\url{https://github.com/Sri-Harsha/scpp}}. The datasheet for \dset is provided in the Supplementary material. The HuggingFace dataset \textbf{Croissant metadata} is available \href{https://huggingface.co/api/datasets/Aman-J/SugarCrepe_pp/croissant}{here}.
\subsection{Author statement} 
In case of violation of rights, the authors will bear all responsibility. We publicly release \dataset under the \textbf{CC-BY-4.0} license.

\subsection{License, Hosting and Maintenance Plan}
We release the dataset publicly under the \textbf{CC-BY-4.0} license on \href{{https://github.com/Sri-Harsha/scpp}}{Github}. The authors of this paper are committed to supporting and maintaining the dataset via our GitHub repository.

%------------------------------------------------------------
\section*{Limitation and Future Work}
In this paper, we evaluate a large set of VLMs on our proposed \dataset. We provide observations based on the results even though it is difficult to provide a comparison between models as the models differ in terms of model architecture, pre-training data content and size, pre-training objectives, etc. Unless we can train separate models for each setting where one of the parameters is kept constant, it is difficult to draw definitive conclusions. Providing guidelines to improve the performance of VLMs on \dataset depends on the observations we draw from the above-mentioned analysis. Thus, in this paper, we limit to identifying one of the major issues with current VLMs, which can probe further research in this direction. Moreover, our dataset can be used to evaluate models to assess their ability to discern lexical alterations from semantic alterations.

\section*{Impact Statement}
This paper presents work whose goal is to advance the field of Machine Learning in general and Language Models research in particular. We discuss several limitations of language models related to the separation of the semantics of an input text from its syntactic and lexical form. In order to build trust-worthy Language Models, it is important to establish that the language models emphasize semantics contained in a sentence rather than the lexical form and syntactic style of the sentence. Our evaluation provides evidence of this problem through two curated datasets and can potentially be impactful for evaluating newer language models and/or inspiring novel solutions to this problem. There are many other potential societal consequences of our work, none of which we feel must be specifically highlighted here.

%------------------------------------------------------------
\section{Datasheet}
\label{sec:datasheet}

\begin{enumerate}[label=Q\arabic*]

\vspace{-0.5em}\begin{figure}[!h]
\subsection{Motivation}\vspace{-1em}
\end{figure}

\item \textbf{For what purpose was the dataset created?} Was there a specific task in mind? Was there a specific gap that needed to be filled? Please provide a description.

\begin{itemize}

\item The \dataset was created to evaluate the sensitivity of vision language models (VLMs) and unimodal language models (ULMs) to semantic and lexical alterations. The \scd dataset consists of (only) one positive and one hard negative caption for each image. Relative to the negative caption, a single positive caption can either have low or high lexical overlap. The original \scd only captures the high overlap case. To evaluate the sensitivity of encoded semantics to lexical alteration, we require an additional positive caption with a different lexical composition. \dset fills this gap by adding an additional positive caption enabling a more thorough assessment of models' abilities to handle semantic content and lexical variation.
\end{itemize}

\item \textbf{Who created the dataset (e.g., which team, research group) and on behalf of which entity (e.g., company, institution, organization)?}

\begin{itemize}
\item  The \dataset is created by the authors of this paper (affiliated with the Faculty of Computer Science, Dalhousie University) to advance our understanding of language models through a new evaluation dataset/task.
\end{itemize}

\item \textbf{Who funded the creation of the dataset?} If there is an associated grant, please provide the name of the grantor and the grant name and number.

\begin{itemize}
\item We acknowledge the support provided by the Faculty of Computer Science, Dalhousie University. Resources used in preparing this research were provided, in part, by the support of the Natural Sciences and Engineering Research Council of Canada
(NSERC), the Province of Ontario, the Government of Canada through Canadian Institute for Advanced Research (CIFAR), ACENET (ace-net.ca), the Digital Research Alliance of Canada (alliancecan.ca) and companies sponsoring the Vector Institute \url{www.vectorinstitute.ai/#partners}.

\end{itemize}

\item \textbf{Any other comments?}

\begin{itemize}
\item No.
\end{itemize}

\vspace{-0.5em}\begin{figure}[!h]
\subsection{Composition}\vspace{-1em}
\end{figure}

\item \textbf{What do the instances that comprise the dataset represent (e.g., documents, photos, people, countries)?} \textit{Are there multiple types of instances (e.g., movies, users, and ratings; people and interactions between them; nodes and edges)? Please provide a description.}

\begin{itemize}
\item The instances from \dataset represent images from MS-COCO \cite{lin2014microsoft} and their associated text captions, negative captions from \scd and newly introduced positive captions.
\end{itemize}

\item \textbf{How many instances are there in total (of each type, if appropriate)?}

\begin{itemize}
\item In total, \dataset consists of 4757 instances. The detailed statistics of the subcategories are provided in \url{https://github.com/Sri-Harsha/scpp}.
\end{itemize}

\item \textbf{Does the dataset contain all possible instances or is it a sample (not necessarily random) of instances from a larger set?} \textit{If the dataset is a sample, then what is the larger set? Is the sample representative of the larger set (e.g., geographic coverage)? If so, please describe how this representativeness was validated/verified. If it is not representative of the larger set, please describe why not (e.g., to cover a more diverse range of instances, because instances were withheld or unavailable).}

\begin{itemize}
\item We included all possible instances from the \scd dataset, except those which are not suitable for our tasks. 
\end{itemize}

\item \textbf{What data does each instance consist of?} \textit{“Raw” data (e.g., unprocessed text or images) or features? In either case, please provide a description.}

\begin{itemize}
\item Each instance of \dataset consists of an image associated with three captions, where two captions describe the image and one caption does not.
\end{itemize}

\item \textbf{Is there a label or target associated with each instance?} \textit{If so, please provide a description.}

\begin{itemize}
\item Each instance in \dset consists of an image and a triplet of captions. The label for a instance is whether each caption in the triplet correctly corresponds to the image or not.
\end{itemize}

\item \textbf{Is any information missing from individual instances?} \textit{If so, please provide a description, explaining why this information is missing (e.g., because it was unavailable). This does not include intentionally removed information, but might include, e.g., redacted text.}

\begin{itemize}
\item No.
\end{itemize}

\item \textbf{Are relationships between individual instances made explicit (e.g., users' movie ratings, social network links)?} \textit{If so, please describe how these relationships are made explicit.}

\begin{itemize}
\item To the best of our knowledge, there is no explicit relationship between the individual instances.
\end{itemize}

\item \textbf{Are there recommended data splits (e.g., training, development/validation, testing)?} \textit{If so, please provide a description of these splits, explaining the rationale behind them.}

\begin{itemize}
\item No, this is only an evaluation dataset.
\end{itemize}

\item \textbf{Are there any errors, sources of noise, or redundancies in the dataset?} \textit{If so, please provide a description.}

\begin{itemize}
\item No, to the best of our knowledge, there are no errors in \dataset. We have done human validation, as described in detail in the paper, to minimize any potential errors.

\end{itemize}

\item \textbf{Is the dataset self-contained, or does it link to or otherwise rely on external resources (e.g., websites, tweets, other datasets)?} \textit{If it links to or relies on external resources, a) are there guarantees that they will exist, and remain constant, over time; b) are there official archival versions of the complete dataset (i.e., including the external resources as they existed at the time the dataset was created); c) are there any restrictions (e.g., licenses, fees) associated with any of the external resources that might apply to a future user? Please provide descriptions of all external resources and any restrictions associated with them, as well as links or other access points, as appropriate.}

\begin{itemize}
\item The images used in our dataset are based on the MS-COCO \cite{lin2014microsoft} dataset, which is freely and publicly available. MS-COCO dataset is released under the Creative Commons Attribution 4.0 license as listed in their website \url{https://cocodataset.org/#termsofuse}.
\end{itemize}

\item \textbf{Does the dataset contain data that might be considered confidential (e.g., data that is protected by legal privilege or by doctor–patient confidentiality, data that includes the content of individuals’ non-public communications)?} \textit{If so, please provide a description.}

\begin{itemize}
\item No, we source part of our dataset, such as image-caption pairs from MS-COCO \cite{lin2014microsoft} and negative captions from SUGARCREPE \cite{hsieh2023sugarcrepe}, both of which are open-source datasets.
\end{itemize}

\item \textbf{Does the dataset contain data that, if viewed directly, might be offensive, insulting, threatening, or might otherwise cause anxiety?} \textit{If so, please describe why.}
\begin{itemize}
\item The authors did not create any content to be explicitly offensive. However, there may be instances that some users may find offensive. Since our \dataset depends on the MS-COCO \cite{lin2014microsoft} and SugarCrepe \cite{hsieh2023sugarcrepe}, we encourage the reader to refer to these dataset's documentation for further details.
\end{itemize}

\item \textbf{Does the dataset relate to people?} \textit{If not, you may skip the remaining questions in this section.}

\begin{itemize}
\item No, the dataset does not relate to people and is not focused on people (although people may appear in the images and descriptions).
\end{itemize}

\item \textbf{Does the dataset identify any subpopulations (e.g., by age, gender)?}
\begin{itemize}
\item We explicitly do not identify any sub-populations.
\end{itemize}

\item \textbf{Is it possible to identify individuals (i.e., one or more natural persons), either directly or indirectly (i.e., in combination with other data) from the dataset?} \textit{If so, please describe how.}
\begin{itemize}
\item Some images might contain identifiable individual faces.
\end{itemize}

\item \textbf{Does the dataset contain data that might be considered sensitive in any way (e.g., data that reveals racial or ethnic origins, sexual orientations, religious beliefs, political opinions or union memberships, or locations; financial or health data; biometric or genetic data; forms of government identification, such as social security numbers; criminal history)?} \textit{If so, please provide a description.}
\begin{itemize}
\item We do not provide any such data in our dataset that may be considered sensitive. All images in our datasets are taken from publicly available datasets.
\end{itemize}

\item \textbf{Any other comments?}

\begin{itemize}
\item No
\end{itemize}

\vspace{-0.5em}\begin{figure}[!h]
\subsection{Collection Process}\vspace{-1em}
\end{figure}

\item \textbf{How was the data associated with each instance acquired?} \textit{Was the data directly observable (e.g., raw text, movie ratings), reported by subjects (e.g., survey responses), or indirectly inferred/derived from other data (e.g., part-of-speech tags, model-based guesses for age or language)? If data was reported by subjects or indirectly inferred/derived from other data, was the data validated/verified? If so, please describe how.}

\begin{itemize}
\item The data associated with each instance
was acquired via our data generation process (see Section 2 in our paper for a detailed description).
\end{itemize}

\item \textbf{What mechanisms or procedures were used to collect the data (e.g., hardware apparatus or sensor, manual human curation, software program, software API)?} \textit{How were these mechanisms or procedures validated?}

\begin{itemize}
\item Please see Section 2 of our paper for a complete description of our data generation and extensive validation process.
\end{itemize}

\item \textbf{If the dataset is a sample from a larger set, what was the sampling strategy (e.g., deterministic, probabilistic with specific sampling probabilities)?}

\begin{itemize}
\item Not applicable.
\end{itemize}

\item \textbf{Who was involved in the data collection process (e.g., students, crowdworkers, contractors) and how were they compensated (e.g., how much were crowdworkers paid)?}

\begin{itemize}
\item The authors of this paper generated the textual content using generative AI as explained in Section 2 of the paper, and manually validated it.

\end{itemize}

\item \textbf{Over what timeframe was the data collected? Does this timeframe match the creation timeframe of the data associated with the instances (e.g., recent crawl of old news articles)?} \textit{If not, please describe the timeframe in which the data associated with the instances was created.}

\begin{itemize}
\item The data was generated and evaluated over the course of approximately four months.
\end{itemize}

\item \textbf{Were any ethical review processes conducted (e.g., by an institutional review board)?} \textit{If so, please provide a description of these review processes, including the outcomes, as well as a link or other access point to any supporting documentation.}

\begin{itemize}
\item We corresponded with the Research Ethics Board (REB) at Dalhousie University. After describing our project in detail, the REB confirmed that our project did not require ethics approval as it did not meet the regulatory definition of human subjects research. Therefore, we did not need to submit a formal application and were allowed to proceed with our research without additional REB review.
\end{itemize}

\item \textbf{Does the dataset relate to people?} \textit{If not, you may skip the remaining questions in this section.}

\begin{itemize}
\item No, the dataset does not relate to people, and is not focused on people (although people may appear in the images and descriptions). 
\end{itemize}

\item \textbf{Did you collect the data from the individuals in question directly, or obtain it via third parties or other sources (e.g., websites)?}

\begin{itemize}
\item Not applicable.
\end{itemize}

\item \textbf{Were the individuals in question notified about the data collection?} \textit{If so, please describe (or show with screenshots or other information) how notice was provided, and provide a link or other access point to, or otherwise reproduce, the exact language of the notification itself.} 

\begin{itemize}
\item Not applicable.
\end{itemize}

\item \textbf{Did the individuals in question consent to the collection and use of their data?} \textit{If so, please describe (or show with screenshots or other information) how consent was requested and provided, and provide a link or other access point to, or otherwise reproduce, the exact language to which the individuals consented.}

\begin{itemize}
\item Not applicable.
\end{itemize}

\item \textbf{If consent was obtained, were the consenting individuals provided with a mechanism to revoke their consent in the future or for certain uses?} \textit{If so, please provide a description, as well as a link or other access point to the mechanism (if appropriate).}
\begin{itemize}
\item Not applicable. 
\end{itemize}

\item \textbf{Has an analysis of the potential impact of the dataset and its use on data subjects (e.g., a data protection impact analysis) been conducted?} \textit{If so, please provide a description of this analysis, including the outcomes, as well as a link or other access point to any supporting documentation.}

\begin{itemize}
\item Not applicable.
\end{itemize}

\item \textbf{Any other comments?}

\begin{itemize}
\item No.
\end{itemize}

\vspace{-0.5em}\begin{figure}[!h]
\subsection{Preprocessing, Cleaning, and/or Labeling}
\vspace{-1em}
\end{figure}

\item \textbf{Was any preprocessing/cleaning/labeling of the data done (e.g., discretization or bucketing, tokenization, part-of-speech tagging, SIFT feature extraction, removal of instances, processing of missing values)?} \textit{If so, please provide a description. If not, you may skip the remainder of the questions in this section.}

\begin{itemize}
\item No preprocessing or labelling was done for creating the scenarios.
\end{itemize}

\item \textbf{Was the “raw” data saved in addition to the preprocessed/cleaned/labeled data (e.g., to support unanticipated future uses)?} \textit{If so, please provide a link or other access point to the “raw” data.}

\begin{itemize}
\item N/A. 
\end{itemize}

\item \textbf{Is the software used to preprocess/clean/label the instances available?} \textit{If so, please provide a link or other access point.}

\begin{itemize}
\item Not applicable
\end{itemize}

\item \textbf{Any other comments?}

\begin{itemize}
\item 
\end{itemize}

\vspace{-0.5em}\begin{figure}[!h]
\subsection{Uses}
\vspace{-1em}
\end{figure}

\item \textbf{Has the dataset been used for any tasks already?} \textit{If so, please provide a description.}

\begin{itemize}
\item No. \dset is a new benchmark.
\end{itemize}

\item \textbf{Is there a repository that links to any or all papers or systems that use the dataset?} \textit{If so, please provide a link or other access point.}

\begin{itemize}
\item To the best of our ability, we will try to maintain links to derivative papers and systems that use our dataset in the \dset GitHub repository (\url{https://github.com/Sri-Harsha/scpp}).
\end{itemize}

\item \textbf{What (other) tasks could the dataset be used for?}

\begin{itemize}
\item The primary use case of our benchmark is to evaluate the sensitivity of VLMs and ULMs to semantic and lexical alterations. While we have not explored this direction in the present work, future work can use this dataset to evaluate any multi-modal system that uses VLMs and ULMs as foundation blocks, such as text-to-image retrieval models, multi-modal chatbots, etc.  
\end{itemize}

\item \textbf{Is there anything about the composition of the dataset or the way it was collected and preprocessed/cleaned/labeled that might impact future uses?} \textit{For example, is there anything that a future user might need to know to avoid uses that could result in unfair treatment of individuals or groups (e.g., stereotyping, quality of service issues) or other undesirable harms (e.g., financial harms, legal risks) If so, please provide a description. Is there anything a future user could do to mitigate these undesirable harms?}
\begin{itemize}
\item Due to the reliance on the MS-COCO \cite{lin2014microsoft} and \scd \cite{hsieh2023sugarcrepe} datasets, \dset may contain offensive material, or biases present in these source datasets. Users of \dset should carefully consider how these limitations may impact their potential use case and exercise discretion in their application of the dataset. 
\end{itemize}

\item \textbf{Are there tasks for which the dataset should not be used?} \textit{If so, please provide a description.}
\begin{itemize}
\item The dataset should be avoided for a
task if the limitations discussed above are unacceptable or potentially problematic for the intended use case.
\end{itemize}

\item \textbf{Any other comments?}

\begin{itemize}
\item No.
\end{itemize}

\vspace{-0.5em}\begin{figure}[!h]
\subsection{Distribution and License}
\vspace{-1em}
\end{figure}

\item \textbf{Will the dataset be distributed to third parties outside of the entity (e.g., company, institution, organization) on behalf of which the dataset was created?} \textit{If so, please provide a description.}

\begin{itemize}
\item Yes, \dataset will be open-sourced and freely available.
\end{itemize}

\item \textbf{How will the dataset be distributed (e.g., tarball on website, API, GitHub)?} \textit{Does the dataset have a digital object identifier (DOI)?}

\begin{itemize}
\item Our datset and code will be made available at the following Github link: \url{https://github.com/Sri-Harsha/scpp}
\end{itemize}

\item \textbf{When will the dataset be distributed?}

\begin{itemize}
\item The dataset will be made available publicly upon publication of this paper.
\end{itemize}

\item \textbf{Will the dataset be distributed under a copyright or other intellectual property (IP) license, and/or under applicable terms of use (ToU)?} \textit{If so, please describe this license and/or ToU, and provide a link or other access point to, or otherwise reproduce, any relevant licensing terms or ToU, as well as any fees associated with these restrictions.}

\begin{itemize}

\item   We release data under the \textbf{CC-BY-4.0} license. 
\item Our code will be released under the \textbf{Apache-2.0} license
\end{itemize}

\item \textbf{Have any third parties imposed IP-based or other restrictions on the data associated with the instances?} \textit{If so, please describe these restrictions, and provide a link or other access point to, or otherwise reproduce, any relevant licensing terms, as well as any fees associated with these restrictions.}

\begin{itemize}
\item The dataset will be released under CC-BY-4.0 license.
\end{itemize}

\item \textbf{Do any export controls or other regulatory restrictions apply to the dataset or to individual instances?} \textit{If so, please describe these restrictions, and provide a link or other access point to, or otherwise reproduce, any supporting documentation.}

\begin{itemize}
\item No.
\end{itemize}

\item \textbf{Any other comments?}
\begin{itemize}
\item No.
\end{itemize}

\vspace{-0.5em}\begin{figure}[!htb]
\subsection{Maintenance}
\vspace{-1em}
\end{figure}

\item \textbf{Who will be supporting/hosting/maintaining the dataset?}

\begin{itemize}
\item The authors will be supporting, hosting and maintaining the dataset and code through GitHub.
\end{itemize}

\item \textbf{How can the owner/curator/manager of the dataset be contacted (e.g., email address)?}

\begin{itemize}
\item The authors can be contacted through their email. Alternatively, an issue can be created on our GitHub repository.
\end{itemize}

\item \textbf{Is there an erratum?} \textit{If so, please provide a link or other access point.}

\begin{itemize}
\item There is no erratum for our initial release. Errata will be documented as future releases on the benchmark website.
\end{itemize}

\item \textbf{Will the dataset be updated (e.g., to correct labeling errors, add new instances, delete instances)?} \textit{If so, please describe how often, by whom, and how updates will be communicated to users (e.g., mailing list, GitHub)?}

\begin{itemize}
\item  \dset will be updated. Updates can be monitored through Github.
\end{itemize}

\item \textbf{If the dataset relates to people, are there applicable limits on the retention of the data associated with the instances (e.g., were individuals in question told that their data would be retained for a fixed period of time and then deleted)?} \textit{If so, please describe these limits and explain how they will be enforced.}

\begin{itemize}
\item NA
\end{itemize}

\item \textbf{Will older versions of the dataset continue to be supported/hosted/maintained?} \textit{If so, please describe how. If not, please describe how its obsolescence will be communicated to users.}

\begin{itemize}
\item We will host older versions in GitHub, in case we release newer versions.
\end{itemize}

\item \textbf{If others want to extend/augment/build on/contribute to the dataset, is there a mechanism for them to do so?} \textit{If so, please provide a description. Will these contributions be validated/verified? If so, please describe how. If not, why not? Is there a process for communicating/distributing these contributions to other users? If so, please provide a description.}

\begin{itemize}
\item Users can extend and build on \dataset as we did for \scd. We do not take responsibility for validating any extension of our work.
\end{itemize}

\item \textbf{Any other comments?}

\begin{itemize}
\item No.
\end{itemize}

\end{enumerate}

\end{document}